\documentclass{article}
\usepackage{multirow}
\usepackage{amsmath}
\usepackage{arydshln}
\usepackage{graphicx}
\usepackage[table]{xcolor}
\usepackage{threeparttable}
\usepackage{tabularray}
\UseTblrLibrary{booktabs}
\usepackage{dashrule}
\usepackage{xspace}
\usepackage{wrapfig}
\usepackage{enumitem}
\usepackage[table]{xcolor}
\usepackage{threeparttable}
\usepackage{tabularray}
\UseTblrLibrary{booktabs}
\definecolor{header}{HTML}{E7EAF3}
\definecolor{headerText}{HTML}{2F3445}
\definecolor{block}{HTML}{F1F3F7}
\definecolor{headerStrong}{HTML}{4F3438}
\definecolor{melody}{HTML}{EAE3F4}
\definecolor{instrument}{HTML}{E1EBF5}
\definecolor{musician}{HTML}{E3EEE2}
\definecolor{genre}{HTML}{F2E8CC}
 \usepackage[preprint]{neurips_2026}

\usepackage[utf8]{inputenc} % allow utf-8 input
\usepackage[T1]{fontenc}    % use 8-bit T1 fonts
\usepackage{hyperref}       % hyperlinks
\usepackage{url}            % simple URL typesetting
\usepackage{booktabs}       % professional-quality tables
\usepackage{amsfonts}       % blackboard math symbols
\usepackage{nicefrac}       % compact symbols for 1/2, etc.
\usepackage{microtype}      % microtypography
\usepackage{xcolor}         % colors
\newcommand{\methodName}{\textsc{TrueMuse}\xspace}

\title{\methodName: A Benchmark for Data Attribution in Text-to-Music Models}

\author{%
  Jiawei Yu \\
  School of Computing \\
  University of Georgia \\
  \texttt{jy40403@uga.edu} \\
  \And
  Jian Liu \\
  School of Computing \\
  University of Georgia \\
  \texttt{jianliu@uga.edu} \\
}

\begin{document}
\maketitle
\begin{abstract}
\vspace{-2mm}
Text-to-music generation models are trained on massive music collections, creating a growing need for data attribution methods that can quantify the contribution of individual training samples. However, existing attribution methods are difficult to rigorously evaluate due to the lack of reliable ground truth, making it challenging to reliably assess their actual effectiveness. To address this gap, we introduce \methodName, a controlled dataset and benchmark for text-to-music data attribution. \methodName is constructed by fine-tuning three diffusion-based text-to-music models on carefully curated attribution samples, whose known inclusion in fine-tuning provides controlled attribution targets for evaluation. The benchmark covers four attribution settings, spanning melodic structure, timbral characteristics, artist-level stylistic signatures, and genre-level shared patterns, and includes 133 attributes, 648 fine-tuned models, and 95,456 generated samples across two prompt types. Using \methodName, we systematically evaluate existing black-box attribution methods along four dimensions: fine-tuning improvement, prompt-type difficulty, multi-task training, and fine-tuning data size. Our results show that attribution remains challenging, with existing methods exhibiting substantial variation across evaluation settings, highlighting the need for more reliable and generalizable attribution methods for text-to-music generation.
\end{abstract}
\vspace{-2mm}
\section{Introduction}
\label{Sec:Introduction}
\vspace{-2mm}
Recent advances in text-to-music generation~\cite{liu2024audioldm, caillon2025live, sunoai2023, udio2024} have achieved breakthrough progress in generating highly realistic, long-form audio, enabling the creation of multi-minute musical pieces with coherent structures and expressive vocals. Such powerful generative capability heavily relies on extensive training over massive music collections, and the commercial value generated by these models is fundamentally derived from learning the creative works of the original artists. This motivates the question of how to credit and compensate creators~\cite{jacques2024protecting} in proportion to each training sample's contribution to a generated output. However, since a single generated piece is simultaneously shaped by multiple training samples with varying degrees of influence~\cite{deng2023computational}, precisely quantifying each sample's contribution remains a fundamental challenge for establishing such a fair royalty distribution system.

To tackle this challenge, researchers have begun exploring data attribution methods for text-to-music generation, though the field remains in its early stages, with existing methods broadly falling into two categories: white-box approaches~\cite{deng2023computational,choi2025largescale} that require access to model parameters, and black-box approaches~\cite{batlle2024towards,barnett2024exploring} that rely on external feature encoders. Specifically, white-box methods operate on model parameters via influence functions~\cite{deng2023computational} or machine unlearning~\cite{choi2025largescale} to estimate training data influence, while black-box methods measure similarity between generated outputs and training samples via external music encoders~\cite{batlle2024towards,barnett2024exploring}.

Despite these efforts, existing attribution methods remain difficult to rigorously evaluate due to the lack of reliable ground truth. White-box methods approximate sample contributions through parameter-based estimation, making it unclear whether the identified samples substantially influenced the generated output. Black-box methods operate on models trained on massive datasets where many samples influence the model parameters simultaneously; this entanglement makes it difficult to isolate the contributions of individual samples to a given generation. Although controlled attribution benchmarks have recently emerged for image generation~\cite{wang2023evaluating}, no comparable benchmark exists for text-to-music generation, where attribution must account for temporal structure, melody, timbre, style, and prompt-conditioned semantics. This gap highlights the need for controlled attribution targets to enable systematic evaluation and support future research.

To fulfill this need, we propose a benchmark for text-to-music data attribution, built upon \methodName, a dataset with known fine-tuning source sets. We construct \methodName by fine-tuning three diffusion-based text-to-music models: AudioLDM2~\cite{liu2024audioldm}, Mustango~\cite{melechovsky2024mustango}, and Stable Audio Open~\cite{evans2025stable}. Inspired by DreamBooth~\cite{ruiz2023dreambooth}, our procedure adapts concept-specific fine-tuning from text-to-image diffusion models to text-to-music generation, binding target musical attributes to selected audio samples. These samples serve as controlled attribution targets because their inclusion in fine-tuning is explicitly known. We design two types of prompts, 15 contextual prompts and up to 28 style prompts per attribute, to generate samples that test attribution methods across different sonic environments and musical styles. \methodName covers four musical attribution sets spanning melodic structure, timbral characteristics, artist-level stylistic signatures, and genre-level shared patterns, across three text-to-music models, comprising 133 attributes, 648 fine-tuned models, and 95,456 generated samples in total. On \methodName, we evaluate similarity-based black-box attribution using four music encoders: MERT-v0~\cite{li2024mert}, MERT-v0-public~\cite{li2024mert}, DAC~\cite{kumar2023high}, and Music2Vec~\cite{li2022map}, along with a contrastive learning-based method~\cite{wang2023evaluating} built on top of these encoders as an enhanced baseline. We assess these methods across four dimensions: attribution improvement from fine-tuned mappers over raw embeddings, attribution difficulty under contextual versus style prompts, the effect of joint multi-task training across melody, musician, instrument, and genre, and the impact of fine-tuning data size by comparing 3 versus 6 clips per attribute. Our results reveal that current attribution methods consistently struggle to achieve reliable performance, highlighting the difficulty of the task and the need for more principled approaches. Our contributions can be summarized as follows:
\begin{itemize}[leftmargin=*]
\item We introduce \methodName, a dataset with known fine-tuning source sets for text-to-music data attribution, constructed via a DreamBooth-based fine-tuning framework across four musical attribution sets and three text-to-music models.
\item We build the first benchmark for text-to-music data attribution on \methodName and conduct a comprehensive evaluation of existing black-box attribution methods across four dimensions, including fine-tuning improvement, prompt type difficulty, multi-task training, and fine-tuning data size.
\item Our results show that attribution remains challenging, with existing methods exhibiting substantial variation across evaluation settings, highlighting the need for more reliable and generalizable attribution methods for text-to-music generation.
\end{itemize}

\vspace{-4mm}
\section{Related Work}
\label{Sec:Related Work}
\vspace{-2mm}
\subsection{Music Generative Models}
\vspace{-2mm}
Early music generation models such as MusicVAE~\cite{roberts2018hierarchical} and MuseGAN~\cite{dong2018musegan} operated over symbolic MIDI representations but were acoustically limited and required separate synthesis pipelines. The field has since shifted toward direct audio synthesis, following three paradigms: diffusion-based models (e.g., AudioLDM2~\cite{liu2024audioldm}, Stable Audio Open~\cite{evans2025stable}, Mustango~\cite{melechovsky2024mustango}) that iteratively denoise acoustic latents conditioned on text; autoregressive transformer models (e.g., MusicGen~\cite{copet2023simple}, Lyria Realtime~\cite{caillon2025live}) that perform sequence modeling over discrete audio tokens; and flow-matching models (e.g., JASCO~\cite{tal2024joint}) that directly map noise to music. Recent works such as YuE~\cite{yuan2025yue} and Jam~\cite{liu2025jam} further explore long-form generation with full vocal performances. On the industry side, Suno~\cite{sunoai2023} and Udio~\cite{udio2024} have demonstrated that text-to-music generation can scale to high-fidelity, full-length compositions, bringing professional-quality music generation to a broad consumer audience.
\vspace{-2mm}
\subsection{Data Attribution}
\vspace{-2mm}
Early data attribution methods were developed primarily for image classification. Influence functions~\cite{koh2017understanding} formalize attribution by approximating the effect of removing a single training example without retraining, and subsequent methods including TracIn~\cite{pruthi2020estimating}, Datamodels~\cite{ilyas2022datamodels}, and TRAK~\cite{park2023trak} improve scalability through alternative gradient estimation strategies. These methods have since been extended to text-to-image diffusion models, with representative works including D-TRAK~\cite{zheng2023intriguing}, Shapley value estimation~\cite{lu2025an}, and Wang et al.~\cite{wang2024data}, which approximate training data influence via machine unlearning. To enable systematic evaluation, Wang et al.~\cite{wang2023evaluating} introduced the first text-to-image attribution benchmark, later adopted by Concept-TRAK~\cite{park2025concept} and A\&D~\cite{liu2026unsupervised}.

While considerable progress has been made in image and text-to-image attribution, the text-to-music domain remains largely unexplored. White-box methods~\cite{deng2023computational,choi2025largescale} assume access to model parameters: Deng et al.~\cite{deng2023computational} apply influence functions to attribute generated music to individual training examples, while Choi et al.~\cite{choi2025largescale} employ machine unlearning in text-to-music diffusion models. Black-box methods~\cite{batlle2024towards,barnett2024exploring} resort to similarity-based retrieval: Barnett et al.~\cite{barnett2024exploring} retrieve training clips via CLAP and CLMR embeddings but impose no control over the generation process, while Batlle-Roca et al.~\cite{batlle2024towards} evaluate similarity metrics on synthetically constructed audio. Across both categories, the lack of reliable ground truth is a fundamental obstacle: without a known source set, assessing the accuracy of an attribution method remains difficult. To address this, we introduce the first evaluation benchmark for text-to-music data attribution, using known fine-tuning source sets as controlled evaluation targets for systematic comparison of attribution methods.
\vspace{-3mm}

\section{Benchmark Details}
\label{Sec:Dataset Curation}
\vspace{-2mm}
\subsection{Dataset Curation}
\vspace{-2mm}
We first revisit the standard training of a text-to-music diffusion model to highlight the challenge of identifying attribution sources. Given a music sample $x$ from a training corpus $\mathcal{X}$ and its text prompt $y$, an audio encoder $\mathcal{E}$ maps $x$ to a latent $z = \mathcal{E}(x)$, a text encoder $g_\phi$ maps $y$ to a conditioning vector $g_\phi(y)$, and a denoising network $\epsilon_\theta$ predicts the noise added to the latent. Together, these components define a parameterized model $\mathcal{M}_{(\theta,\phi)}$. Its pre-trained instance $\mathcal{M}_{(\theta_{\mathrm{pre}}, \phi_{\mathrm{pre}})}$ is obtained by minimizing the standard diffusion objective:
{\small
\begin{equation}
(\theta_{\mathrm{pre}}, \phi_{\mathrm{pre}}) = \operatorname*{arg\,min}_{\theta,\phi} \, \mathbb{E}_{x \sim \mathcal{X},\, y,\, \epsilon \sim \mathcal{N}(0,\mathbf{I}),\, t}\!\left[\|\epsilon - \epsilon_\theta(z_t,\, t,\, g_\phi(y))\|_2^2\right],
\end{equation}
}
where $t$ is a uniformly sampled timestep, $\epsilon$ is the Gaussian noise added to the latent $z$ to produce the noisy latent $z_t$ at $t$, and $\epsilon_\theta(z_t,\, t,\, g_\phi(y))$ is the network's prediction of this noise. While this enables high-quality music generation, a model trained on a large-scale corpus is influenced by a vast number of training samples simultaneously, making it difficult to identify the contribution of an individual sample to a given generation and to evaluate attribution methods against reliable ground truth.

\begin{figure}[t]
    \centering
    \includegraphics[width=1\linewidth]{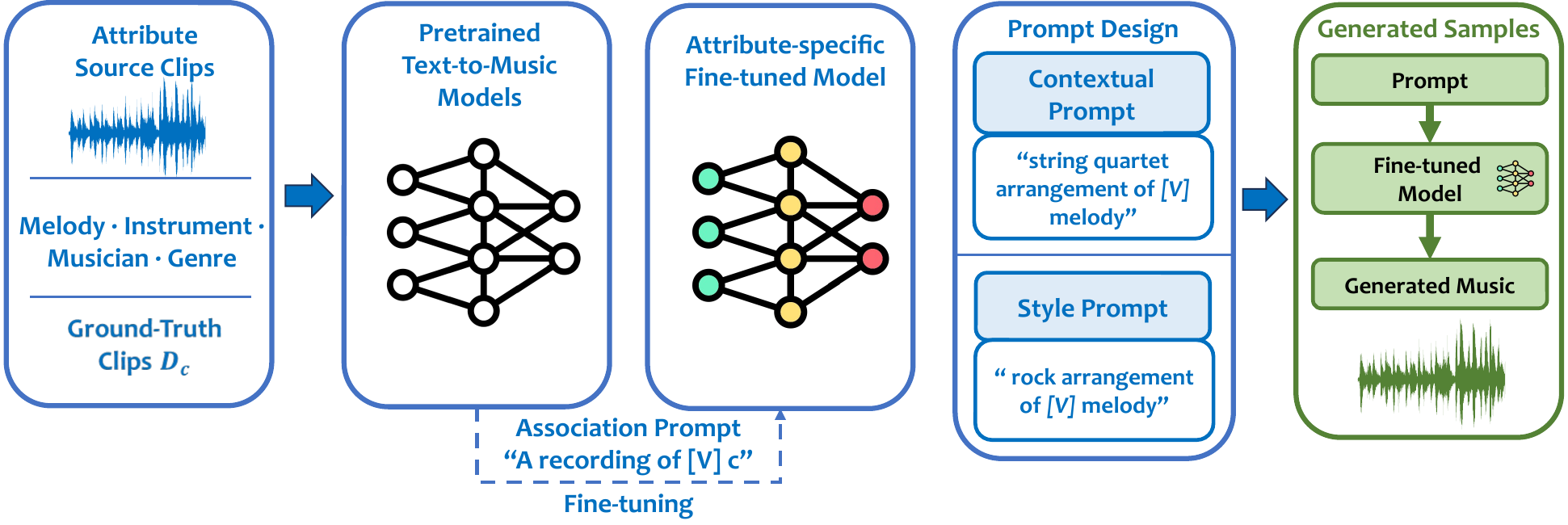}
    \vspace{-4mm}
    \caption{Overview of the \methodName construction pipeline. For each target attribute, selected source clips are used to fine-tune a pre-trained text-to-music model with an association prompt, and the resulting attribute-specific model generates samples from contextual and style prompts with known fine-tuning sources.}
    \label{fig:overview}
    \vspace{-4mm}
\end{figure}

To address this, inspired by DreamBooth~\cite{ruiz2023dreambooth}, we fine-tune text-to-music models to establish controlled relationships between selected source samples and generated outputs, as illustrated in Figure~\ref{fig:overview}. Specifically, let $c$ denote a target musical attribute (e.g., a melody, instrument, musician, or genre), and let $\mathcal{D}_c = \{x_1^c, x_2^c, \dots, x_N^c\}$ be a small set of clips of the attribute $c$. Starting from the pre-trained model $\mathcal{M}_{(\theta_{\mathrm{pre}},\phi_{\mathrm{pre}})}$, we fine-tune the denoising network $\epsilon_\theta$ on $\mathcal{D}_c$ using the attribute prompt $y_c$ = ``a recording of $\texttt{[V]}$ $c$'', where \texttt{[V]} is a rare identifier (e.g., \texttt{sks}) that binds to $\mathcal{D}_c$. This association aims to steer outputs $\hat{x}$ generated by the resulting model conditioned on $y_c$ toward $\mathcal{D}_c$. The fine-tuning objective is:
{\small
\begin{equation}
\theta_c^* = \operatorname*{arg\,min}_{\theta} \, \mathbb{E}_{x^c \sim \mathcal{D}_c,\, y_c,\, \epsilon \sim \mathcal{N}(0,\mathbf{I}),\, t}\!\left[\|\epsilon - \epsilon_{\theta}(z^c_t,\, t,\, g_{\phi_{\mathrm{pre}}}(y_c))\|_2^2\right],
\end{equation}
}
where $z^c = \mathcal{E}(x^c)$, $z^c_t$ is the noisy latent of $z^c$ at $t$, and $g_{\phi_{\mathrm{pre}}}$ is kept frozen throughout. The source set $\mathcal{D}_c$ used to fine-tune the resulting model $\mathcal{M}_{(\theta_c^*,\, \phi_{\mathrm{pre}})}$ is explicitly known and serves as the evaluation target for attribution methods.

As shown in Table~\ref{Tab:Total Number}, we design four types of attribution sets for \methodName to evaluate attribution methods across diverse musical characteristics, ranging from fine-grained acoustic features to high-level stylistic patterns.
\begin{itemize}[leftmargin=*]
    \item \textbf{Melody-Centric}: Each attribute corresponds to a specific melodic theme, instantiated as a single short clip carrying a well-defined pitch contour (e.g., a folk tune or a classical theme). We use $50$ such clips, randomly sampled from the Free Music Archive~\cite{defferrard2016fma} (FMA), as the attributes. An effective attribution method should be able to capture melodic structure that is invariant to changes in arrangement and instrumentation.
    \item \textbf{Instrument-Centric}: Each attribute corresponds to a specific instrument (e.g., acoustic guitar, piano), represented by a small set of clips in which that instrument is clearly dominant. We use $25$ instruments covering diverse timbral categories as the attributes, with clips collected from YouTube under fair use for non-commercial research purposes. An effective attribution method should be able to capture timbral characteristics that are invariant to changes in musical context.
    \item \textbf{Musician-Centric}: Each attribute corresponds to a specific artist (e.g., a jazz pianist or a folk singer), represented by a small set of clips drawn from that artist's own recordings. We use $50$ artists, randomly sampled from FMA, as the attributes. An effective attribution method should be able to capture stylistic signatures that persist across varied compositions.
    \item \textbf{Genre-Centric}: Each attribute corresponds to a genre (e.g., jazz, rock), represented by clips drawn from multiple different artists within that genre. We use the $8$ most frequently occurring genres in FMA as the attributes. An effective attribution method should be able to capture stylistic patterns shared across different artists rather than tied to any individual's identity.
\end{itemize}

% =========================
% Table
% =========================
\begin{table*}[t]
\centering
\caption{Dataset statistics across four attribution sets and three text-to-music models.}
\vspace{+2mm}
\label{Tab:Total Number}
\scriptsize
\begin{threeparttable}
\begin{tblr}{
  width = \linewidth,
  colspec = {
    X[1.15,c,m]
    X[1.20,c,m]
    X[0.95,c,m]
    X[0.78,c,m]
    X[0.92,c,m]
    X[0.92,c,m]
    X[0.72,c,m]
    X[0.74,c,m]
  },
  rows = {abovesep=0.8pt, belowsep=0.8pt},
  column{1-3} = {bg=block},
  column{8} = {bg=block},
  hline{1} = {1-8}{1pt},
  hline{2} = {1-8}{0.8pt},
  hline{6} = {1-8}{0.5pt},
  hline{8} = {1-8}{0.5pt},
  hline{11} = {1-8}{0.5pt},
  hline{14} = {1-8}{0.5pt},
  hline{17} = {1-8}{0.8pt},
  hline{18} = {1-8}{1pt},
  row{1} = {bg=headerStrong, fg=white, font=\bfseries},
  row{17} = {bg=headerStrong, fg=white, font=\bfseries},
  row{11,14} = {abovesep=3pt},
}
% ===== header =====
\SetCell[c=3]{c,bg=headerStrong,fg=white,font=\bfseries} Property & & 
& \SetCell{bg=headerStrong,fg=white,font=\bfseries} Melody
& \SetCell{bg=headerStrong,fg=white,font=\bfseries} Instrument
& \SetCell{bg=headerStrong,fg=white,font=\bfseries} Musician
& \SetCell{bg=headerStrong,fg=white,font=\bfseries} Genre
& \SetCell{bg=headerStrong,fg=white,font=\bfseries} Total \\
% ===== top stats =====
\SetCell[c=3]{c,bg=block} Attributes & & 
& 50 & 25 & 50 & 8 & 133 \\
\SetCell[c=3]{c,bg=block} Total finetuning clips & & 
& 50 & 225 & 450 & 72 & 797 \\
\SetCell[c=3]{c,bg=block} Clips per model & & 
& 1 & 3/6 & 3/6 & 3/6 & -- \\
\SetCell[c=3]{c,bg=block} Total fine-tuned models & & 
& 150 & 150 & 300 & 48 & 648 \\
% ===== prompts =====
\SetCell[r=2]{c,bg=block} Prompts
& \SetCell[c=2]{c,bg=block} Contextual & 
& 15 & 15 & 15 & 15 & -- \\
& \SetCell[c=2]{c,bg=block} Style\tnote{1} & 
& 28/28/8 & 28/28/8 & 28/28/8 & 28/28/28 & -- \\
% ===== samples: AudioLDM2 =====
\SetCell[r=9]{c,bg=block} Samples
& \SetCell[r=3]{c,bg=block} AudioLDM2
& Contextual
& 3,000 & 3,000 & 6,000 & 960 & 12,960 \\
&
& Style
& 5,600 & 5,600 & 11,200 & 1,792 & 24,192 \\
&
& \textbf{Total}
& \textbf{8,600} & \textbf{8,600} & \textbf{17,200} & \textbf{2,752} & \textbf{37,152} \\
% ===== samples: Mustango =====
&
\SetCell[r=3]{c,bg=block} Mustango
& Contextual
& 3,000 & 3,000 & 6,000 & 960 & 12,960 \\
&
& Style
& 5,600 & 5,600 & 11,200 & 1,792 & 24,192 \\
&
& \textbf{Total}
& \textbf{8,600} & \textbf{8,600} & \textbf{17,200} & \textbf{2,752} & \textbf{37,152} \\
% ===== samples: Stable Audio =====
&
\SetCell[r=3]{c,bg=block} Stable Audio Open
& Contextual
& 3,000 & 3,000 & 6,000 & 960 & 12,960 \\
&
& Style
& 1,600 & 1,600 & 3,200 & 1,792 & 8,192 \\
&
& \textbf{Total}
& \textbf{4,600} & \textbf{4,600} & \textbf{9,200} & \textbf{2,752} & \textbf{21,152} \\
% ===== total =====
\SetCell[c=3]{c,bg=headerStrong,fg=white,font=\bfseries} Total samples & &
& \textbf{21,800}
& \textbf{21,800}
& \textbf{43,600}
& \textbf{8,256}
& \textbf{95,456} \\
\end{tblr}
\begin{tablenotes}[flushleft]
\scriptsize
\item[1]
Reported as AudioLDM2 / Mustango / Stable Audio Open. Stable Audio Open uses 8 style prompts for Melody, Instrument, and Musician due to its limited caption diversity; Genre uses 28 across all models.
\end{tablenotes}
\end{threeparttable}
\vspace{-6mm}
\end{table*}

These four attribution sets are instantiated across three diffusion-based text-to-music models: AudioLDM2~\cite{liu2024audioldm}, Mustango~\cite{melechovsky2024mustango}, and Stable Audio Open~\cite{evans2025stable}, where each attribute is fine-tuned independently on each model. In total, this yields 648 fine-tuned models across all attribution sets and models.

For the generation of samples, we design two types of prompts: \textit{contextual prompts}, which vary the musical scene and accompaniment context to test whether attribution methods remain effective when attributes appear in different sonic environments, and \textit{style prompts}, which vary the musical genre to test whether attribution methods remain effective when attributes are presented in different styles. All prompts are produced using Claude~\cite{anthropic2026claude}: for each text-to-music model, we provide Claude with official example prompts from that model's demonstration website and issue two separate instructions, one for contextual prompts and one for style prompts, asking it to generate variations that follow the same style and structure as the examples. The official examples used for each model and examples of generated prompts are provided in Appendix~\ref{App:ClaudePrompts}. This procedure ensures that the generated prompts align with each model's expected input distribution, thereby maintaining generation quality.

For Stable Audio Open, style prompts are restricted to the 8 most frequently occurring genre tags in its training corpus due to its limited training caption diversity; this restriction applies to Melody, Instrument, and Musician, but not to Genre-Centric where all three models use 28 style prompts. For each fine-tuned model, 4 samples are generated per prompt to reduce the effect of randomness in the generation process. In total, our dataset comprises 95,456 generated samples across four attribution sets and three models.

\vspace{-2mm}
\subsubsection{Melody-Centric}
\vspace{-2mm}
This set tests whether attribution methods can identify a specific melody when the model re-arranges it with different instruments or in a different style. We use $1$ clip per attribute because a melody is uniquely determined by its pitch contour and adding more clips would introduce variability from other tracks that no longer correspond to the same melody. The $50$ melodies are randomly sampled from the Free Music Archive~\cite{defferrard2016fma} (FMA), and we fine-tune with $y_c$ = ``a recording of $\texttt{[V]}$ melody''. Contextual prompts vary the sonic environment (e.g., ``a string quartet arrangement of $\texttt{[V]}$ melody, warm and intimate''), and style prompts vary the genre (e.g., ``a rock arrangement of $\texttt{[V]}$ melody'').
\vspace{-2mm}
\subsubsection{Instrument-Centric}
\vspace{-2mm}
This set tests whether attribution methods can identify a specific instrument across different melodies, accompaniments, and styles. We use $3$ and $6$ clips per attribute because, unlike a melody, an instrument's timbre varies with playing technique, register, and articulation, and a single clip captures only a narrow slice of this range; the two configurations let us study how attribution performance depends on this coverage. The $25$ instruments cover diverse timbral categories (e.g., acoustic guitar, drums), with clips collected from YouTube under fair use for non-commercial academic research; we use these clips solely for fine-tuning and do not redistribute the original audio. We fine-tune with $y_c$ = ``a recording of $\texttt{[V]}$ [instrument]''. Contextual prompts take the form ``a catchy beat with $\texttt{[V]}$ [instrument] in the mix'', and style prompts take the form ``a rock song with $\texttt{[V]}$ [instrument]''.
\vspace{-2mm}
\subsubsection{Musician-Centric}
\vspace{-2mm}
This set tests whether attribution methods can identify a specific artist across different instrumentations and genres, which is the most common case in real-world disputes such as copyright and royalty allocation. We use the same $3$ and $6$ clip configurations as the Instrument-Centric set, with each clip drawn from a different track of the same artist so that the fine-tuned model captures recurring stylistic patterns rather than memorizing one song. The $50$ musicians are randomly sampled from FMA, and we fine-tune with $y_c$ = ``a recording of $\texttt{[V]}$ musician''. Contextual prompts take the form ``a $\texttt{[V]}$ musician singing softly over a music box and muted strings'', and style prompts take the form ``a rock song by $\texttt{[V]}$ musician''.
\vspace{-2mm}
\subsubsection{Genre-Centric}
\vspace{-2mm}
This set tests whether attribution methods can identify a genre as a pattern shared across artists, rather than misattributing it to any single artist contained in the training clips. We use the same $3$ and $6$ clip configurations, with the clips for each genre drawn from $6$ different musicians so that no individual artist signature dominates the fine-tuning signal. The $8$ genres are the most frequently occurring ones in FMA, and we fine-tune with $y_c$ = ``a recording of $\texttt{[V]}$ genre''. Contextual prompts follow the same form as previous sets (e.g., ``a warm acoustic melody in $\texttt{[V]}$ [genre] with fingerpicked guitar''). Since the genre is the attribution target, style prompts instead fix the genre and vary the instrument (e.g., ``a guitar performance in $\texttt{[V]}$ [genre]'').
\vspace{-2mm}
\subsection{Evaluating for Attribution}
\label{Sec:Evaluating}
\vspace{-2mm}
Following prior work on black-box attribution~\cite{batlle2024towards,barnett2024exploring}, we formulate attribution as a retrieval task. For each generated sample $\hat{x}$ produced by a model fine-tuned on attribute $c$, its corresponding $\mathcal{D}_c$ serves as the designated source set to be retrieved. At evaluation stage, we use $\hat{x}$ as the query and match it against a gallery containing clips from $\mathcal{D}_c$.

We evaluate two methods under this retrieval protocol. The first, which we take as our standard baseline following prior work~\cite{batlle2024towards,barnett2024exploring}, uses a frozen music encoder $F_{\text{base}}$ to encode both $\hat{x}$ and $\mathcal{D}_c$, and performs retrieval directly on these embeddings by ranking candidates with cosine similarity. The second, adapted from Wang et al.~\cite{wang2023evaluating}, improves the baseline by introducing the \textit{mapper}, a linear projection head applied to the output embeddings of the frozen music encoder $F_{\text{base}}$, trained with a contrastive loss on generated samples and their ground-truth clips to produce embeddings better suited for attribution. Specifically, for a frozen music encoder $F_{\text{base}}$, two mappers are applied, $h$ with weight matrix $W_h$ for clips in $\mathcal{D}_c$ and $\tilde{h}$ with weight matrix $W_{\tilde{h}}$ for $\hat{x}$, and trained jointly with a symmetric NT-Xent contrastive loss~\cite{chen2020simple}:
{\small
\begin{equation}
\mathcal{L}_{\text{cont}} = -\frac{1}{2}\left(
\log \frac{\exp(\tilde{h}(F_{\text{base}}(\hat{x}))^\top h(F_{\text{base}}(x^{c+})) / \beta)}{\sum_{j} \exp(\tilde{h}(F_{\text{base}}(\hat{x}))^\top h(F_{\text{base}}(x^c_j)) / \beta)}
+ \log \frac{\exp(h(F_{\text{base}}(x^{c+}))^\top \tilde{h}(F_{\text{base}}(\hat{x})) / \beta)}{\sum_{j} \exp(h(F_{\text{base}}(x^{c+}))^\top \tilde{h}(F_{\text{base}}(\hat{x}_j)) / \beta)}
\right),
\end{equation}
}
where $x^{c+} \in \mathcal{D}_c$ is a ground-truth clip for $\hat{x}$, $x^c_j$ and $\hat{x}_j$ are negatives drawn from other attributes in the batch, and $\beta$ is a temperature. During training, only the two mappers $h$ and $\tilde{h}$ are optimized while $F_{\text{base}}$ remains frozen. We also add a Frobenius norm regularization on the mapper weights, giving the final training objective $\mathcal{L}_{\text{attr}} = \mathcal{L}_{\text{cont}} + \lambda \left( \|W_h\|_F + \|W_{\tilde{h}}\|_F \right)$, where $\lambda$ controls the regularization strength.

It is important to note that we focus on black-box methods because existing white-box approaches~\cite{deng2023computational,choi2025largescale} do not release source code and are often tied to specific text-to-music models and training data, making faithful reproduction and fair comparison difficult. Black-box methods, by contrast, require no access to text-to-music model parameters, gradients, or large-scale training procedures, enabling consistent evaluation across methods.
\vspace{-4mm}
\section{Experiments}
\label{Sec:Experiments}
\vspace{-2mm}
We evaluate both the standard baseline and the mapper method on four attribution sets: Melody-Centric, Instrument-Centric, Musician-Centric, and Genre-Centric, across three text-to-music generators: AudioLDM2, Mustango, and Stable Audio Open. Details for these generators can be found in Appendix~\ref{sec:generatordetail}. Within each attribution set, attributes are partitioned into train, validation, and test splits with a $70/15/15$ ratio; mappers are trained on train-split attributes, and all reported results are on test-split attributes. For Instrument-Centric, Musician-Centric, and Genre-Centric, we report results using 3-clips per attribute in the main paper and include the 6-clip version in the Appendix~\ref{sec:6clipsettings}. We resample all ground-truth clips and generated samples to 16~kHz to ensure consistency during retrieval. We evaluate four music encoders: MERT-v0~\cite{li2024mert}, MERT-v0-public~\cite{li2024mert}, Music2Vec-v1~\cite{li2022map}, and DAC-16k~\cite{kumar2023high}. Details of these encoders can be found in Appendix~\ref{app:encoders}.

We use Recall@$k$ as our evaluation metric, defined as the fraction of ground-truth clips from $\mathcal{D}_c$ retrieved within the top-$k$ candidates, averaged over all generated queries. To simulate a realistic scenario in which ground-truth clips are accompanied by a large amount of unrelated music, we augment the retrieval gallery with $228{,}000$ ten-second distractor clips from JamendoMaxCaps~\cite{roy2025jamendomaxcaps} in addition to the ground-truth clips from the test-split attributes. We report Recall@$10$ in the main paper, defined as the fraction of ground-truth clips from $\mathcal{D}_c$ retrieved within the top-$10$ candidates, and provide additional metrics such as Recall@$1$, Recall@$5$, Recall@$100$, and Mean Average Precision (MAP) in the Appendix~\ref{sec:differentmetics}. We adopt cosine similarity as the similarity measurement for all experiments. We introduce the finetuning settings of text-to-music models in Appendix~\ref{sec:detailsfort2m} and the training settings of mappers in Appendix~\ref{sec:detailsformappers}.

We first compare the standard method with the mapper method across four attribution sets, four encoders, and three generators to establish a comprehensive baseline. We then conduct three experiments using mappers to comprehensively evaluate how their performance varies across different scenarios: (i) we examine how prompt type, including contextual-only, style-only, and mixed prompts, affects mapper training and attribution performance; (ii) we investigate whether mappers trained on all four attribution sets improve attribution over those trained on a single attribution set; and (iii) we study how the number of fine-tuning clips per attribute ($3$ versus $6$) affects attribution. By default, we use samples generated from both contextual and style prompts for training, unless otherwise specified.

\begin{figure}[t]
    \centering
    \includegraphics[width=1\linewidth]{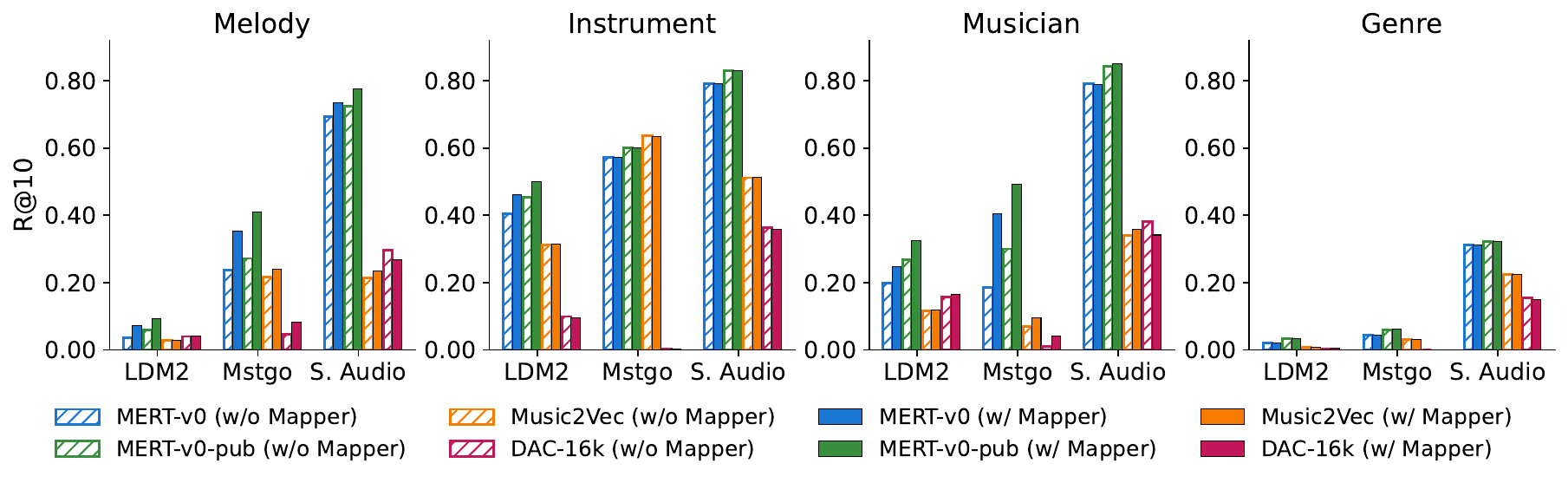}
    \vspace{-6mm}
    \caption{Attribution R@10 performance of the standard method and the mapper-enhanced method. The generator names are abbreviated as LDM2 (AudioLDM2), Mstgo (Mustango), and S. Audio (Stable Audio Open).}
    \label{fig:Exp1}
    \vspace{-6mm}
\end{figure}

\vspace{-4mm}
\subsection{General Attribution Performance}
\vspace{-2mm}

\noindent\textbf{Attribution Performance Across Encoders.}
As shown in Figure~\ref{fig:Exp1}, the four encoders produce noticeably different attribution performance. MERT-v0-public is the strongest at R@10 of $0.397$, followed by MERT-v0 at $0.357$, Music2Vec-v1 at $0.226$, and DAC-16k at $0.129$, all averaged over all attribution sets and generators. DAC-16k, a neural codec trained to reconstruct audio waveforms, falls behind MERT by a wide margin, suggesting that attribution depends more on musical semantic understanding than on acoustic reconstruction fidelity. The 4.0 percentage-point gap between MERT-v0 and MERT-v0-public, which share the same architecture but differ only in pre-training corpus, further indicates that pre-training data alone can meaningfully shift attribution performance.

\noindent\textit{Key Finding:} [Encoder choice has a substantial impact on attribution performance. Models trained with music-specific self-supervised objectives substantially outperform codec-based encoders, and even within the same architecture, pre-training data has a meaningful effect. The gap between DAC-16k and MERT reflects a fundamental difference in training objective: reconstruction versus music understanding.]

\noindent\textbf{Attribution Performance Across Generators.}
Figure~\ref{fig:Exp1} reveals that the three generators produce very different attribution difficulty. Stable Audio Open is the easiest to attribute, reaching R@10 of $0.487$, compared to $0.205$ for Mustango and $0.140$ for AudioLDM2. Such difference may be attributed to Stable Audio Open's tendency to produce outputs that closely resemble its fine-tuning examples, possibly due to its constrained prompt format, which leaves little room for variation across generations, or an architectural tendency to memorize, both of which push generated samples closer to the training clips.

\noindent\textit{Key Finding:} [Attribution difficulty varies substantially across generators and is governed by how distinctively each generator reproduces its fine-tuning clips. Stable Audio Open's outputs may remain closer to the fine-tuning clips, making them much easier to retrieve than those of Mustango or AudioLDM2.]

\noindent\textbf{Attribution Performance Across Attribution Sets.}
Attribution difficulty also varies sharply across the four attribution sets, as can be seen in Figure~\ref{fig:Exp1}. Instrument-Centric is the easiest at R@10 of $0.465$, followed by Musician-Centric at $0.305$, Melody-Centric at $0.239$, and Genre-Centric at $0.101$. Such ordering likely reflects the acoustic specificity of each attribute: instruments leave a clear timbral fingerprint that encoders can detect directly, whereas melody is defined by a pitch contour that can be reproduced under many different instrumentations, and genre is a high-level category with no single corresponding acoustic feature, requiring the encoder to integrate multiple characteristics across clips.

\noindent\textit{Key Finding:} [Attribution difficulty reflects the acoustic complexity of the target attribute. Fine-grained acoustic features such as instrument timbre are easier to retrieve than abstract musical concepts such as genre. The difficulty ordering from Instrument-Centric to Genre-Centric mirrors the increasing level of abstraction, from a concrete physical property of sound to a high-level stylistic category.]

\noindent\textbf{Attribution Performance with Mappers.}
Training mappers on top of the frozen encoders improves attribution mainly for MERT-based encoders, as shown in Figure~\ref{fig:Exp1}, with MERT-v0 improving from $0.357$ to $0.400$ and MERT-v0-public from $0.397$ to $0.441$. Gains are largest on Mustango, at $0.084$ for both MERT-v0 and MERT-v0-public, and minimal on Stable Audio Open, at most $0.015$, where the encoder space is already near saturation. Music2Vec-v1 and DAC-16k show little benefit. Such pattern may be attributed to the mapper being effective when the encoder already encodes attribution-relevant information but generated samples and ground-truth clips remain misaligned in the embedding space, allowing the mapper to learn a small linear correction that closes the gap. When the encoder lacks discriminative signal or the retrieval space is already near saturation, a linear projection cannot recover information the encoder did not capture.

\noindent\textit{Key Finding:} [Learned mappers provide meaningful gains only when the encoder already captures attribution-relevant information but generated samples and ground-truth clips remain misaligned. Gains being concentrated on Mustango and almost absent on Stable Audio Open are consistent with this explanation: Stable Audio Open's outputs are already close to the fine-tuning clips, leaving little room for the mapper to improve.]

\begin{figure}[t]
  \centering
  \includegraphics[width=1\linewidth]{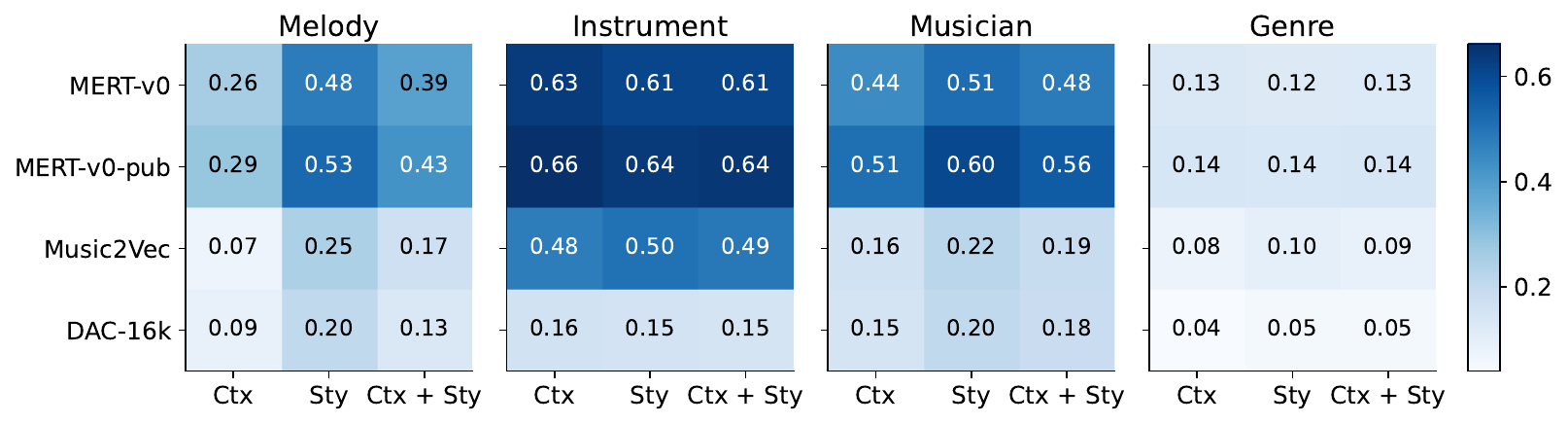}
  \vspace{-6mm}
  \caption{Attribution performance under different prompt types. R@10 is reported for mappers trained on contextual prompts only (Ctx), style prompts only (Sty), and contextual plus style prompts (Ctx + Sty).}
  \vspace{-6mm}
  \label{fig:exp2}
\end{figure}

\vspace{-5mm}
\subsection{Attribution Performance Under Different Prompt Types}
\vspace{-2mm}
\noindent\textbf{Style Prompts Outperform Contextual Prompts.}
As shown in Figure~\ref{fig:exp2}, style prompts outperform contextual prompts on average across all encoders. Averaged over attribution sets, MERT-v0-public reaches $0.478$ under style training compared to $0.400$ under contextual, and MERT-v0 reaches $0.430$ compared to $0.365$. Music2Vec-v1 and DAC-16k show the same trend at lower absolute levels. The gain is concentrated on Melody-Centric and Musician-Centric, while Instrument-Centric and Genre-Centric remain nearly unchanged. Combining contextual and style prompts does not consistently improve over style-only training, often yielding performance between the two single-prompt settings. This suggests that style prompts provide a cleaner and more attribute-focused training signal for the mapper.

\noindent\textit{Key Finding:} [Style prompts outperform contextual prompts on average, with gains concentrated on attribution sets where the target attribute is easily underspecified. Combining prompt types provides no consistent benefit over style-only training.]

\noindent\textbf{Melody-Centric Is Most Sensitive to Prompt Type.}
The effect of prompt type varies substantially across attribution sets. Melody-Centric is the most sensitive, with R@10 increasing from $0.178$ under contextual prompts to $0.365$ under style prompts. Musician-Centric also improves from $0.315$ to $0.383$, while Instrument-Centric and Genre-Centric remain largely flat across contextual, style, and mixed settings. This suggests that contextual prompts often leave the target melody underconstrained, whereas style prompts anchor generation more tightly to the target melodic structure.

\noindent\textit{Key Finding:} [Prompt type matters most when the target attribute can be easily underspecified by the prompt, as in Melody-Centric, and has much less effect when the attribute already dominates the acoustic output.]

\begin{figure}[t]
    \centering
    \includegraphics[width=1\linewidth]{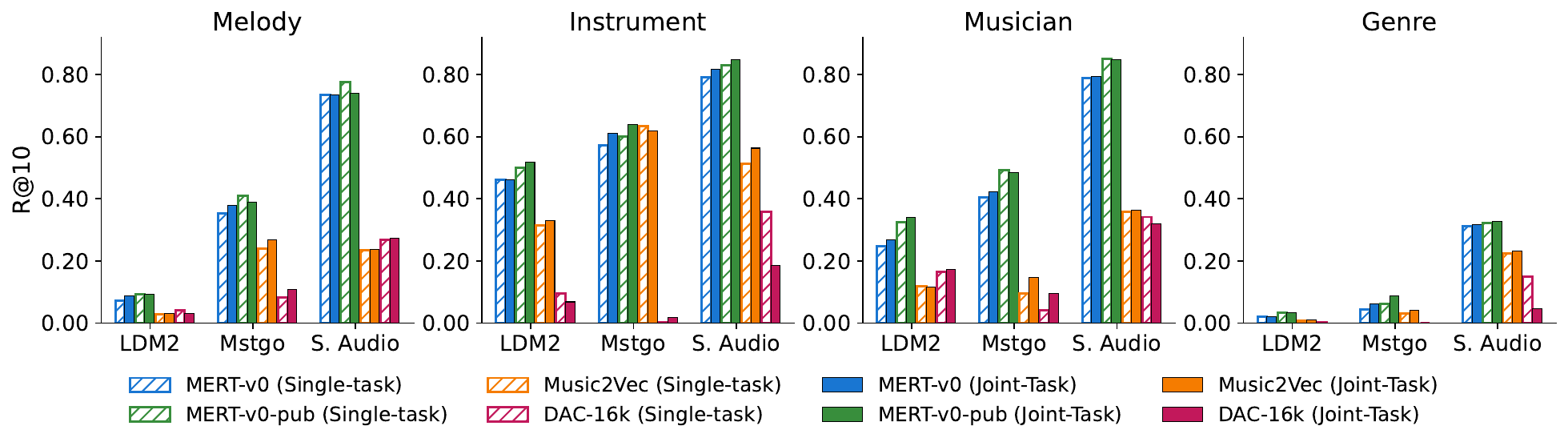}
    \vspace{-6mm}
    \caption{Attribution R@10 comparing a mapper trained on a single attribution set (Single-task) against one trained jointly on all four attribution sets (Joint-task). The generator names are abbreviated as LDM2 (AudioLDM2), Mstgo (Mustango), and S. Audio (Stable Audio Open).}
    \vspace{-4mm}
    \label{fig:exp3}
\end{figure}

\vspace{-4mm}
\subsection{Single-Task versus Joint-Task Attribution}
\vspace{-2mm}
\noindent\textbf{Joint Training Performs Comparably to Single-Task Training.}
As shown in Figure~\ref{fig:exp3}, the difference between single-task and joint-task training is small across all attribution sets. Musician-Centric shows the largest gain under joint training, from $0.352$ to $0.364$, while Melody-Centric and Instrument-Centric remain essentially unchanged, and Genre-Centric shows a slight drop from $0.101$ to $0.098$. No encoder benefits substantially from joint training, and the overall average changes by only $0.003$. Such negligible difference may be attributed to the four attribution sets targeting fundamentally different acoustic characteristics that occupy largely disjoint regions of the embedding space, leaving little shared structure for joint training to exploit.

\noindent\textit{Key Finding:} [A single jointly trained mapper performs comparably to four separately trained mappers. The four attribution sets cover largely orthogonal dimensions of musical content, and a single mapper can learn attribution-relevant projections for all four without one task degrading the others.]

\begin{figure}[t]
    \centering
    \includegraphics[width=0.85\linewidth]{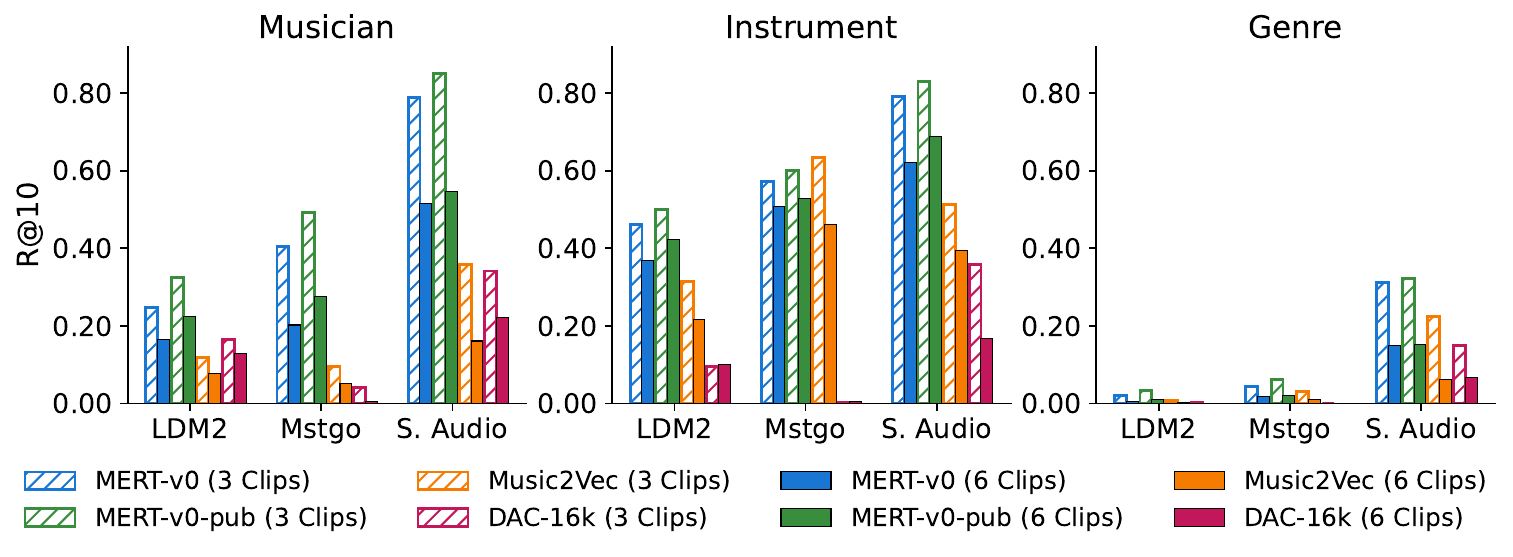}
    \vspace{-3mm}
    \caption{Attribution R@10 performance comparing mapper training with $3$ versus $6$ ground-truth clips per attribute. The generator names are abbreviated as LDM2 (AudioLDM2), Mstgo (Mustango), and S. Audio (Stable Audio Open).}
    \vspace{-6mm}
    \label{fig:exp5}
\end{figure}

\vspace{-4mm}
\subsection{Attribution Performance with Varying Clip Counts}
\vspace{-2mm}

\noindent\textbf{Attribution Degrades Consistently with More Clips.}
As shown in Figure~\ref{fig:exp5}, increasing the number of fine-tuning clips from $3$ to $6$ consistently reduces R@10 across the Instrument-Centric, Musician-Centric, and Genre-Centric settings, as well as across encoders and generators. Averaged over all conditions, R@10 drops from $0.309$ to $0.210$. Musician-Centric shows the largest drop, from $0.352$ to $0.214$, followed by Instrument-Centric from $0.473$ to $0.374$, and Genre-Centric from $0.101$ to $0.041$. The drop is consistent across encoders: MERT-v0-public drops by $0.128$, MERT-v0 by $0.121$, Music2Vec-v1 by $0.096$, and DAC-16k by $0.051$. The effect is most pronounced for Stable Audio Open, falling by $0.174$, compared to $0.075$ for Mustango and $0.048$ for AudioLDM2. Increasing the number of fine-tuning clips expands the candidate source set and may make individual clips harder to retrieve. The effect is most pronounced for Musician-Centric, where six clips drawn from different tracks introduce broader stylistic variation, further diffusing the model's outputs across the embedding space.

\noindent\textit{Key Finding:} [More fine-tuning clips consistently reduces attribution performance. With more clips, the fine-tuned model is exposed to a broader range of acoustic realizations, producing more diverse outputs that spread further across the embedding space and become harder to retrieve against a fixed set of ground-truth clips.]

\vspace{-5mm}
\section{Discussion and Limitations}
\label{Sec:Discussion}
\vspace{-3mm}
\methodName provides a controlled benchmark with known fine-tuning source sets for systematically evaluating text-to-music attribution methods. Our results show that source retrieval remains challenging and varies substantially across musical attributes and generators. As an initial benchmark, \methodName focuses on a representative set of musical attributes, text-to-music generators, and prompt conditions; future extensions may further expand the coverage of models, musical concepts, and generation scenarios. We hope this benchmark encourages more rigorous study of attribution in text-to-music systems and supports the development of methods that are more robust, interpretable, and suitable for real-world credit assignment.
\vspace{-2mm}

\bibliographystyle{plain} 
\bibliography{references} 

@inproceedings{park2023trak,
author = {Park, Sung Min and Georgiev, Kristian and Ilyas, Andrew and Leclerc, Guillaume and M\k{a}dry, Aleksander},
title = {TRAK: attributing model behavior at scale},
year = {2023},
publisher = {JMLR.org},
booktitle = {Proceedings of the 40th International Conference on Machine Learning},
articleno = {1128},
numpages = {40},
location = {Honolulu, Hawaii, USA},
series = {ICML'23}
}

@inproceedings{koh2017understanding,
  title={Understanding black-box predictions via influence functions},
  author={Koh, Pang Wei and Liang, Percy},
  booktitle={International conference on machine learning},
  pages={1885--1894},
  year={2017},
  organization={PMLR}
}

@inproceedings{li2022map,
  title={MAP-Music2Vec: A Simple and Effective Baseline for Self-Supervised Music Audio Representation Learning},
  author={Li, Yizhi and Yuan, Ruibin and Zhang, Ge and MA, Yinghao and Lin, Chenghua and Chen, Xingran and Ragni, Anton and Yin, Hanzhi and Hu, Zhijie and He, Haoyu and others},
  booktitle={Ismir 2022 Hybrid Conference},
  year={2022}
}

@article{jacques2024protecting,
  title={Protecting human creativity in AI-generated music with the introduction of an AI-royalty fund},
  author={Jacques, Sabine and Flynn, Mathew},
  journal={GRUR International},
  volume={73},
  number={12},
  pages={1137--1149},
  year={2024},
  publisher={Oxford University Press UK}
}

@article{pruthi2020estimating,
  title={Estimating training data influence by tracing gradient descent},
  author={Pruthi, Garima and Liu, Frederick and Kale, Satyen and Sundararajan, Mukund},
  journal={Advances in Neural Information Processing Systems},
  volume={33},
  pages={19920--19930},
  year={2020}
}

@inproceedings{
caillon2025live,
title={Live Music Models},
author={Antoine Caillon and Brian McWilliams and Cassie Tarakajian and Ian Simon and Ilaria Manco and Jesse Engel and Noah Constant and Yunpeng Li and Timo I. Denk and Alberto Lalama and Andrea Agostinelli and Cheng-Zhi Anna Huang and Ethan Manilow and George Brower and Hakan Erdogan and Heidi Lei and Itai Rolnick and Ivan Grishchenko and Manu Orsini and Matej Kastelic and Mauricio Zuluaga and Mauro Verzetti and Michael Dooley and Ondrej Skopek and Rafael Ferrer and Zal{\'a}n Borsos and Aaron van den Oord and Douglas Eck and Eli Collins and Jason Michael Baldridge and Tom Hume and Chris Donahue and Kehang Han and Adam Roberts},
booktitle={The Thirty-ninth Annual Conference on Neural Information Processing Systems Creative AI Track: Humanity},
year={2025},
url={https://openreview.net/forum?id=SB3XbPgvpl}
}

@inproceedings{roy2025jamendomaxcaps,
  title={Jamendomaxcaps: A large scale music-caption dataset with imputed metadata},
  author={Roy, Abhinaba and Liu, Renhang and Lu, Tongyu and Herremans, Dorien},
  booktitle={2025 International Joint Conference on Neural Networks (IJCNN)},
  pages={1--8},
  year={2025},
  organization={IEEE}
}

@article{tal2024joint,
  title={Joint audio and symbolic conditioning for temporally controlled text-to-music generation},
  author={Tal, Or and Ziv, Alon and Gat, Itai and Kreuk, Felix and Adi, Yossi},
  journal={arXiv preprint arXiv:2406.10970},
  year={2024}
}

@misc{sunoai2023,
  author = {{Suno, Inc.}},
  title = {{Suno: AI Music Generator}},
  howpublished = {\url{https://suno.com}},
  month={[Accessed: 2026-4-19]}
}

@misc{udio2024,
  author = {{Udio, Inc.}},
  title = {{Udio: AI Music Generator}},
  howpublished = {\url{https://www.udio.com}},
  month={[Accessed: 2026-4-19]}
}

@article{copet2023simple,
  title={Simple and controllable music generation},
  author={Copet, Jade and Kreuk, Felix and Gat, Itai and Remez, Tal and Kant, David and Synnaeve, Gabriel and Adi, Yossi and D{\'e}fossez, Alexandre},
  journal={Advances in neural information processing systems},
  volume={36},
  pages={47704--47720},
  year={2023}
}

@inproceedings{dong2018musegan,
  title={Musegan: Multi-track sequential generative adversarial networks for symbolic music generation and accompaniment},
  author={Dong, Hao-Wen and Hsiao, Wen-Yi and Yang, Li-Chia and Yang, Yi-Hsuan},
  booktitle={Proceedings of the AAAI conference on artificial intelligence},
  volume={32},
  number={1},
  year={2018}
}

@inproceedings{roberts2018hierarchical,
  title={A hierarchical latent vector model for learning long-term structure in music},
  author={Roberts, Adam and Engel, Jesse and Raffel, Colin and Hawthorne, Curtis and Eck, Douglas},
  booktitle={International conference on machine learning},
  pages={4364--4373},
  year={2018},
  organization={PMLR}
}

@article{liu2025jam,
  title={JAM: A Tiny Flow-based Song Generator with Fine-grained Controllability and Aesthetic Alignment},
  author={Liu, Renhang and Hung, Chia-Yu and Majumder, Navonil and Gautreaux, Taylor and Bagherzadeh, Amir Ali and Li, Chuan and Herremans, Dorien and Poria, Soujanya},
  journal={arXiv preprint arXiv:2507.20880},
  year={2025}
}

@article{deng2023computational,
  title={Computational copyright: Towards a royalty model for music generative ai},
  author={Deng, Junwei and Zhang, Shiyuan and Ma, Jiaqi},
  journal={arXiv preprint arXiv:2312.06646},
  year={2023}
}

@article{yuan2025yue,
  title={Yue: Scaling open foundation models for long-form music generation},
  author={Yuan, Ruibin and Lin, Hanfeng and Guo, Shuyue and Zhang, Ge and Pan, Jiahao and Zang, Yongyi and Liu, Haohe and Liang, Yiming and Ma, Wenye and Du, Xingjian and others},
  journal={arXiv preprint arXiv:2503.08638},
  year={2025}
}

@inproceedings{chen2020simple,
  title={A simple framework for contrastive learning of visual representations},
  author={Chen, Ting and Kornblith, Simon and Norouzi, Mohammad and Hinton, Geoffrey},
  booktitle={International conference on machine learning},
  pages={1597--1607},
  year={2020},
  organization={PmLR}
}

@article{kumar2023high,
  title={High-fidelity audio compression with improved rvqgan},
  author={Kumar, Rithesh and Seetharaman, Prem and Luebs, Alejandro and Kumar, Ishaan and Kumar, Kundan},
  journal={Advances in Neural Information Processing Systems},
  volume={36},
  pages={27980--27993},
  year={2023}
}

@inproceedings{
li2024mert,
title={{MERT}: Acoustic Music Understanding Model with Large-Scale Self-supervised Training},
author={Yizhi LI and Ruibin Yuan and Ge Zhang and Yinghao Ma and Xingran Chen and Hanzhi Yin and Chenghao Xiao and Chenghua Lin and Anton Ragni and Emmanouil Benetos and Norbert Gyenge and Roger Dannenberg and Ruibo Liu and Wenhu Chen and Gus Xia and Yemin Shi and Wenhao Huang and Zili Wang and Yike Guo and Jie Fu},
booktitle={The Twelfth International Conference on Learning Representations},
year={2024},
url={https://openreview.net/forum?id=w3YZ9MSlBu}
}

@article{ilyas2022datamodels,
  title={Datamodels: Predicting predictions from training data},
  author={Ilyas, Andrew and Park, Sung Min and Engstrom, Logan and Leclerc, Guillaume and Madry, Aleksander},
  journal={arXiv preprint arXiv:2202.00622},
  year={2022}
}

@article{wang2024data,
  title={Data attribution for text-to-image models by unlearning synthesized images},
  author={Wang, Sheng-Yu and Hertzmann, Aaron and Efros, Alexei A and Zhu, Jun-Yan and Zhang, Richard},
  journal={Advances in Neural Information Processing Systems},
  volume={37},
  pages={4235--4266},
  year={2024}
}

@inproceedings{
choi2025largescale,
title={Large-Scale Training Data Attribution for Music Generative Models via Unlearning},
author={Woosung Choi and Junghyun Koo and Kin Wai Cheuk and Joan Serr{\`a} and Marco A. Mart{\'\i}nez-Ram{\'\i}rez and Yukara Ikemiya and Naoki Murata and Yuhta Takida and Wei-Hsiang Liao and Yuki Mitsufuji},
booktitle={The Thirty-ninth Annual Conference on Neural Information Processing Systems Creative AI Track: Humanity},
year={2025},
url={https://openreview.net/forum?id=qj3ps8lNIf}
}

@inproceedings{wang2023evaluating,
  title={Evaluating data attribution for text-to-image models},
  author={Wang, Sheng-Yu and Efros, Alexei A and Zhu, Jun-Yan and Zhang, Richard},
  booktitle={Proceedings of the IEEE/CVF International Conference on Computer Vision},
  pages={7192--7203},
  year={2023}
}

@inproceedings{ruiz2023dreambooth,
  title={Dreambooth: Fine tuning text-to-image diffusion models for subject-driven generation},
  author={Ruiz, Nataniel and Li, Yuanzhen and Jampani, Varun and Pritch, Yael and Rubinstein, Michael and Aberman, Kfir},
  booktitle={Proceedings of the IEEE/CVF conference on computer vision and pattern recognition},
  pages={22500--22510},
  year={2023}
}

@misc{anthropic2026claude,
  author = {Anthropic},
  title = {Claude},
  year = {2026},
  url = {https://claude.ai/},
  note = {Version 4.6 Sonnet, Accessed: 2026-04-17}
}

@article{defferrard2016fma,
  title={FMA: A dataset for music analysis},
  author={Defferrard, Micha{\"e}l and Benzi, Kirell and Vandergheynst, Pierre and Bresson, Xavier},
  journal={arXiv preprint arXiv:1612.01840},
  year={2016}
}

@inproceedings{evans2025stable,
  title={Stable audio open},
  author={Evans, Zach and Parker, Julian D and Carr, CJ and Zukowski, Zack and Taylor, Josiah and Pons, Jordi},
  booktitle={ICASSP 2025-2025 IEEE International Conference on Acoustics, Speech and Signal Processing (ICASSP)},
  pages={1--5},
  year={2025},
  organization={IEEE}
}

@inproceedings{melechovsky2024mustango,
  title={Mustango: Toward controllable text-to-music generation},
  author={Melechovsky, Jan and Guo, Zixun and Ghosal, Deepanway and Majumder, Navonil and Herremans, Dorien and Poria, Soujanya},
  booktitle={Proceedings of the 2024 Conference of the North American Chapter of the Association for Computational Linguistics: Human Language Technologies (Volume 1: Long Papers)},
  pages={8293--8316},
  year={2024}
}

@article{liu2024audioldm,
  title={Audioldm 2: Learning holistic audio generation with self-supervised pretraining},
  author={Liu, Haohe and Yuan, Yi and Liu, Xubo and Mei, Xinhao and Kong, Qiuqiang and Tian, Qiao and Wang, Yuping and Wang, Wenwu and Wang, Yuxuan and Plumbley, Mark D},
  journal={IEEE/ACM Transactions on Audio, Speech, and Language Processing},
  volume={32},
  pages={2871--2883},
  year={2024},
  publisher={IEEE}
}

@article{park2025concept,
  title={Concept-TRAK: Understanding how diffusion models learn concepts through concept-level attribution},
  author={Park, Yonghyun and Lai, Chieh-Hsin and Hayakawa, Satoshi and Takida, Yuhta and Murata, Naoki and Liao, Wei-Hsiang and Choi, Woosung and Cheuk, Kin Wai and Koo, Junghyun and Mitsufuji, Yuki},
  journal={arXiv preprint arXiv:2507.06547},
  year={2025}
}

@article{barnett2024exploring,
  title={Exploring musical roots: Applying audio embeddings to empower influence attribution for a generative music model},
  author={Barnett, Julia and Garcia, Hugo Flores and Pardo, Bryan},
  journal={arXiv preprint arXiv:2401.14542},
  year={2024}
}

@article{batlle2024towards,
  title={Towards assessing data replication in music generation with music similarity metrics on raw audio},
  author={Batlle-Roca, Roser and Liao, Wei-Hsiang and Serra, Xavier and Mitsufuji, Yuki and G{\'o}mez, Emilia},
  journal={arXiv preprint arXiv:2407.14364},
  year={2024}
}

@article{liu2026unsupervised,
  title={Unsupervised Synthetic Image Attribution: Alignment and Disentanglement},
  author={Liu, Zongfang and Chen, Guangyi and Sun, Boyang and Liu, Tongliang and Zhang, Kun},
  journal={arXiv preprint arXiv:2601.22663},
  year={2026}
}

@article{zheng2023intriguing,
  title={Intriguing properties of data attribution on diffusion models},
  author={Zheng, Xiaosen and Pang, Tianyu and Du, Chao and Jiang, Jing and Lin, Min},
  journal={arXiv preprint arXiv:2311.00500},
  year={2023}
}

@inproceedings{
lu2025an,
title={An Efficient Framework for Crediting Data Contributors of Diffusion Models},
author={MingYu Lu and Chris Lin and Chanwoo Kim and Su-In Lee},
booktitle={The Thirteenth International Conference on Learning Representations},
year={2025},
url={https://openreview.net/forum?id=9EqQC2ct4H}
}
%%%%%%%%%%%%%%%%%%%%%%%%%%%%%%%%%%%%%%%%%%%%%%%%%%%%%%%%%%%%

\appendix

\section{Technical Appendices and Supplementary Material}

\subsection{Prompt Construction with Claude}
\label{App:ClaudePrompts}

For each text-to-music model and each attribution set, we collect one or a few official example prompts from the model's demonstration website and provide Claude~\cite{anthropic2026claude} with two separate short instructions, one for each prompt type. The contextual prompt instruction is along the lines of: ``Following the style of the example prompt(s) below, generate $15$ contextual prompts that vary the musical scene and accompaniment around the attribute $\texttt{[V]}$ $c$ while keeping the attribute itself fixed.'' The style prompt instruction is along the lines of: ``Following the style of the example prompt(s) below, generate $28$ style prompts of the form `a [genre] song by/with $\texttt{[V]}$ $c$' that vary the musical genre while keeping the attribute fixed.'' The prompts returned by Claude are used as-is during sample generation. We list the official example prompts used as references and the full set of generated prompts for one representative case (Musician-Centric on AudioLDM2) below; the prompts for the remaining (model, attribution set) combinations are included in our released benchmark.

\subsubsection{Official Example Prompts}
\begin{itemize}
    \item \textbf{AudioLDM2}: \texttt{A catchy trap beat with EDM Synthesizers in the mix.}
    \item \textbf{Mustango}: \texttt{This techno song features a synth lead playing the main melody ... The chord sequence is Gm, A7, Eb, Bb, C, F, Gm. The tempo is 128.0 beats per minute. The key is G minor.}; \texttt{The song is an instrumental ... a classical guitar playing a lilting melody ... The chord sequence is Gm, F6, Ebm. The time signature is 4/4. This song is in Adagio. The key is G minor.}
    \item \textbf{Stable Audio Open}: \texttt{128 BPM tech house drum loop}
\end{itemize}

\subsubsection{Generated Contextual Prompts (Musician-Centric, AudioLDM2)}
\begin{itemize}\small
    \item A $\texttt{[V]}$ musician performing a heartfelt set over gentle piano chords.
    \item A $\texttt{[V]}$ musician laying down a late-night groove with warm Rhodes keys.
    \item A $\texttt{[V]}$ musician playing smooth downtempo on a rooftop at sunset.
    \item A $\texttt{[V]}$ musician performing a reflective piece with soft piano on a rainy day.
    \item A $\texttt{[V]}$ musician playing a light bossa nova with acoustic guitar in a morning cafe.
    \item A $\texttt{[V]}$ musician leading a powerful stadium anthem with driving drums.
    \item A $\texttt{[V]}$ musician rolling out a dusty groove with slide guitar.
    \item A $\texttt{[V]}$ musician dropping a thumping nightclub banger with heavy bass.
    \item A $\texttt{[V]}$ musician singing softly over a music box and muted strings.
    \item A $\texttt{[V]}$ musician strumming an easy folk tune by a campfire with group harmonies.
    \item A $\texttt{[V]}$ musician building an epic cinematic theme with swelling orchestra.
    \item A $\texttt{[V]}$ musician riding a gritty boom-bap loop with sharp hi-hats.
    \item A $\texttt{[V]}$ musician floating across airy synth pads and wind chimes.
    \item A $\texttt{[V]}$ musician firing up a funky jam with horns and hand percussion.
    \item A $\texttt{[V]}$ musician delivering an emotional ballad over delicate fingerpicked guitar.
\end{itemize}

\subsubsection{Generated Style Prompts (Musician-Centric, AudioLDM2)}
All style prompts follow the form ``a [genre] song by $\texttt{[V]}$ musician'', where [genre] ranges over the following $28$ genre tags: pop, rock, hip hop, R\&B, jazz, classical, blues, country, electronic, soul, funk, disco, reggae, metal, folk, Latin, indie, dance, gospel, trap, punk, K-pop, lo-fi, alternative, house, techno, reggaeton, and ambient.

\subsection{Generator Details}
\label{sec:generatordetail}
Table~\ref{tab:generatorsversion} summarizes the three text-to-music generators included in \methodName.
All three are latent diffusion models conditioned on text, but differ in architecture and conditioning scope.
\begin{itemize}
    \item \textbf{AudioLDM2}~\cite{liu2024audioldm} employs a U-Net denoiser conditioned on a joint audio-language embedding produced by CLAP and GPT-2, enabling flexible cross-modal text-to-audio generation.
    \item \textbf{Mustango}~\cite{melechovsky2024mustango} extends this paradigm with an additional music-domain conditioning branch that injects beat and chord predictions, allowing the model to follow music-theoretic constraints specified in the prompt.
    \item \textbf{Stable Audio Open}~\cite{evans2025stable} adopts a similar U-Net backbone conditioned on T5 text embeddings, and is the largest of the three at 1.1B parameters.
\end{itemize}
Together, these three generators cover a spectrum of model scales (712M--1.4B parameters), conditioning strategies, and training corpora, making them a diverse testbed for evaluating attribution across different generative pipelines.

Both AudioLDM2 and Mustango produce 10-second clips at 16 kHz, while Stable Audio Open produces 10-second clips at 44.1 kHz. We resample samples generated by Stable Audio Open to 16 kHz to ensure consistency. For AudioLDM2 we use 200 DDIM steps; for Mustango we use 100 diffusion steps with a classifier-free guidance scale of 3.0; for Stable Audio Open we use 200 steps with a guidance scale of 3.0. Each finetuned model generates 4 waveforms per prompt, yielding 4 samples per (attribute, prompt) pair.

\begin{table}[h]
\centering
\caption{Specifications for the three text-to-music generators used in \methodName.}
\label{tab:generatorsversion}
\footnotesize
\begin{threeparttable}
\begin{tblr}{
  width = \linewidth,
  colspec = {
    X[1.8,c,m]
    X[0.8,c,m]
    X[2.2,c,m]
    X[0.8,c,m]
  },
  rows = {abovesep=1.2pt, belowsep=1.2pt},
  hline{1} = {1-4}{1pt},
  hline{2} = {1-4}{0.8pt},
  hline{5} = {1-4}{0.8pt},
  hline{6} = {1-4}{1pt},
  row{1} = {bg=headerStrong, fg=white, font=\bfseries},
  row{5} = {bg=block},
  row{3} = {bg=block},
}
Generator & Year & Version / Checkpoint & \#Params \\
AudioLDM2~\cite{liu2024audioldm} & 2024 & cvssp/audioldm2 & 712M \\
Mustango~\cite{melechovsky2024mustango} & 2024 & declare-lab/mustango & 1.4B \\
Stable Audio Open~\cite{evans2025stable} & 2025 & stabilityai/stable-audio-open-1.0 & 1.1B \\
\end{tblr}
\end{threeparttable}
\end{table}

\subsection{Attribution Performance on 6-Clip Setting}
\label{sec:6clipsettings}
\begin{figure}[h]
    \centering
    \includegraphics[width=1\linewidth]{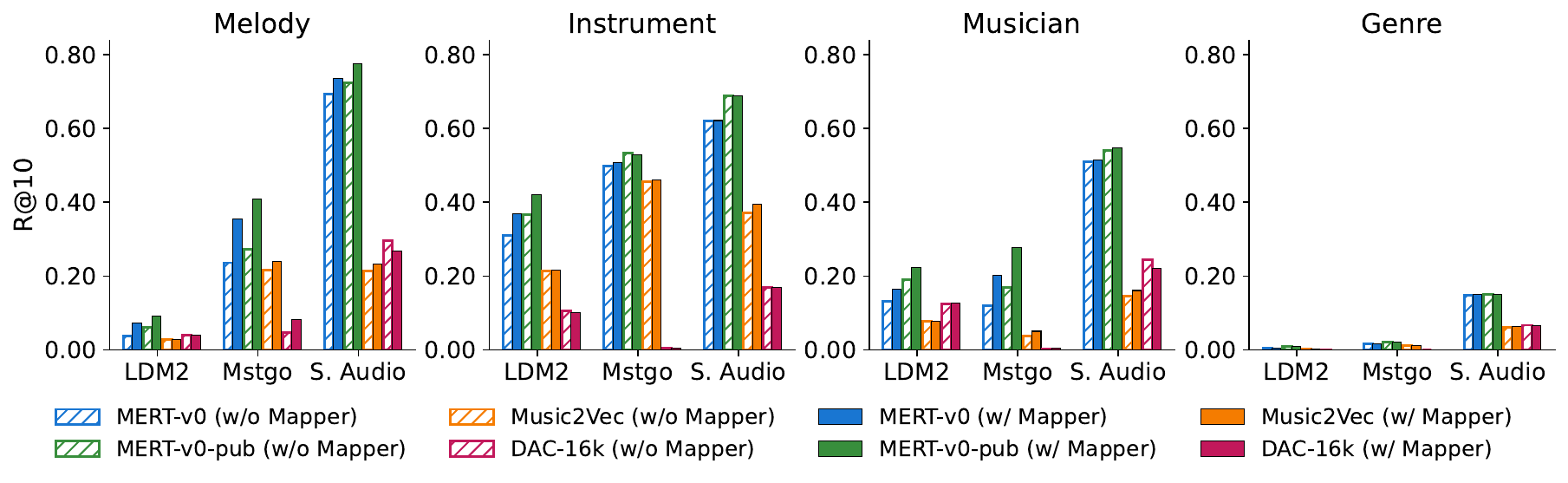}
    \caption{Attribution R@10 performance of the standard method and the mapper-enhanced method, using $6$ clips per attribute. The generator names are abbreviated as LDM2 (AudioLDM2), Mstgo (Mustango), and S. Audio (Stable Audio Open).}
    \label{fig:main-result-6clips}
\end{figure}

\begin{figure}[h]
  \centering
  \includegraphics[width=1\linewidth]{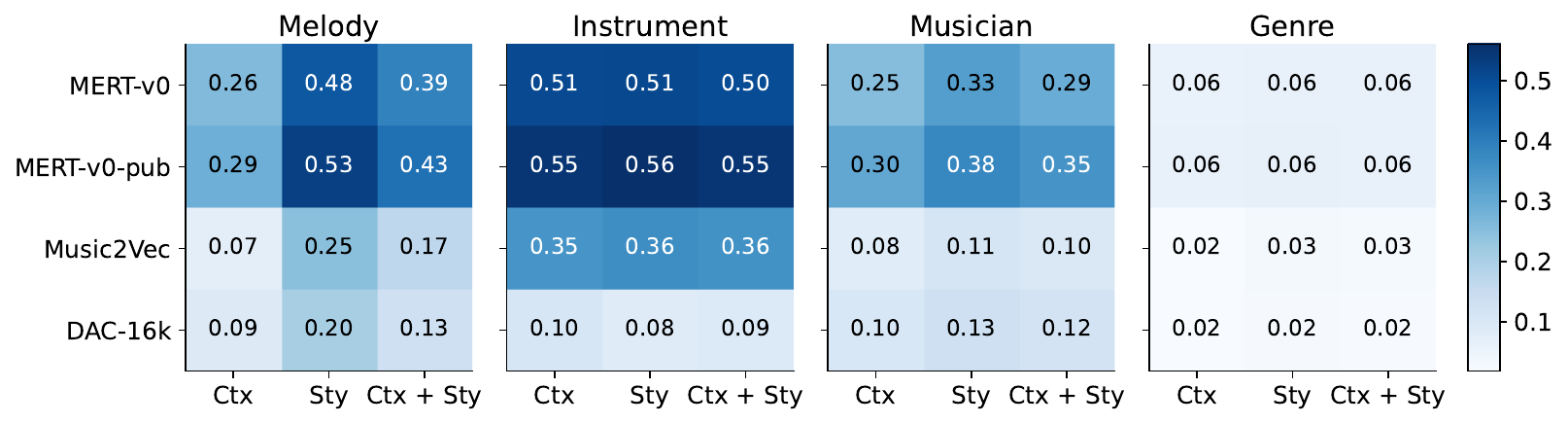}
    \caption{Attribution performance under different prompt types, using $6$ clips per attribute. R@10 is reported for mappers trained on contextual prompts only (Ctx), style prompts only (Sty), and contextual plus style prompts (Ctx + Sty).}
  \label{fig:multiprompt-6clips}
\end{figure}

\begin{figure}[h]
    \centering
    \includegraphics[width=1\linewidth]{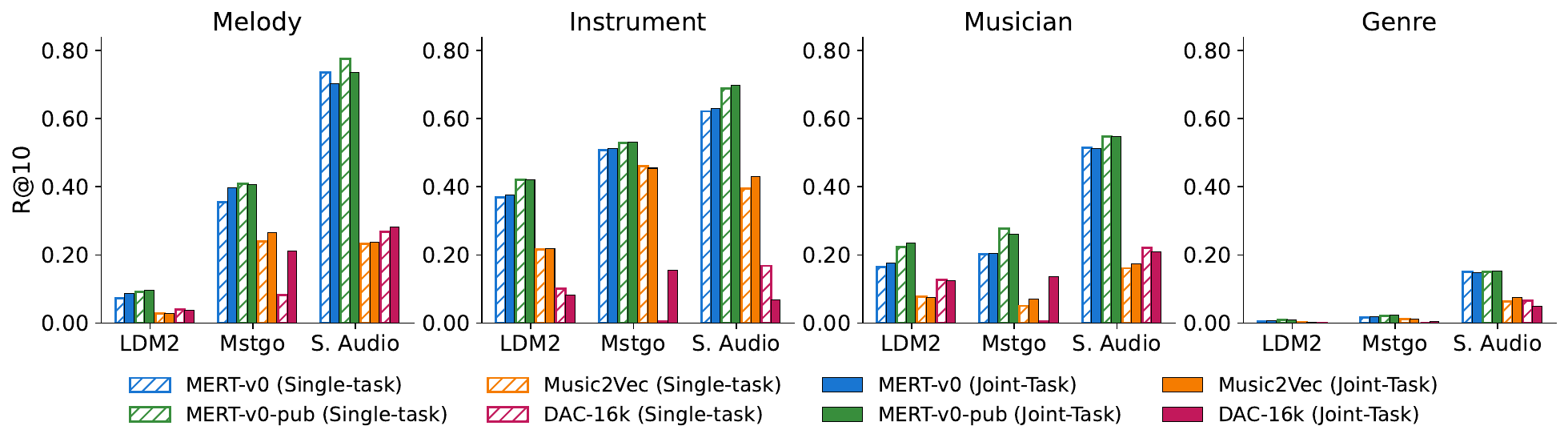}
    \caption{Attribution R@10 comparing a mapper trained on a single attribution set against one trained jointly on all four attribution sets, using $6$ clips per attribute. The generator names are abbreviated as LDM2 (AudioLDM2), Mstgo (Mustango), and S. Audio (Stable Audio Open).}
    \label{fig:multitask-6clips}
\end{figure}

We replicate the main experiments from Section~\ref{Sec:Experiments} using the 6-clip setting for Instrument-Centric, Musician-Centric, and Genre-Centric; Melody-Centric remains in the 1-clip setting. As discussed in the main text, performance consistently degrades relative to the 3-clip setting across encoders, generators, and the three non-melody attribution sets. The relative ordering across encoders and generators remains unchanged: MERT-based encoders continue to outperform Music2Vec-v1 and DAC-16k, Stable Audio Open remains the easiest generator to attribute, and learned mappers still provide the largest gains on Mustango for MERT-based encoders. The prompt type ordering also holds, with style prompts continuing to outperform contextual prompts in most conditions. Results are shown in Figures~\ref{fig:main-result-6clips},~\ref{fig:multiprompt-6clips}, and~\ref{fig:multitask-6clips}.

\subsection{Music Encoder Details}
\label{app:encoders}
We evaluate four music encoders spanning different pretraining paradigms and representation strategies.
MERT-v0 and MERT-v0-pub~\cite{li2024mert} share the same RoBERTa-style transformer backbone (12 layers, $D{=}768$) pretrained via masked acoustic modelling with an acoustic teacher signal; MERT-v0-pub is trained exclusively on publicly licensed data, whereas MERT-v0 additionally uses a private corpus.
Music2Vec-v1~\cite{li2022map} is a data2vec-based model pretrained on music audio with a self-distillation objective. DAC-16k~\cite{kumar2023high} is a residual vector quantization codec with 12 codebooks of size 1024. Table~\ref{tab:encoders} summarizes the raw output format, aggregation procedure, and final embedding dimension for each encoder. All encoders operate on 16\,kHz mono audio and are kept frozen throughout all experiments.

\begin{table*}[h]
\centering
\caption{Summary of music encoders used in \methodName.}
\vspace{+2mm}
\label{tab:encoders}
\footnotesize
\begin{threeparttable}
\begin{tblr}{
  width = \linewidth,
  colspec = {
    X[1.8,c,m]
    X[2.0,c,m]
    X[3.2,c,m]
    X[0.9,c,m]
  },
  rows = {abovesep=1.2pt, belowsep=1.2pt},
  column{1} = {bg=block},
  hline{1} = {1-4}{1pt},
  hline{2} = {1-4}{0.8pt},
  hline{4} = {1-4}{0.5pt},
  hline{5} = {1-4}{0.5pt},
  hline{6} = {1-4}{1pt},
  row{1} = {bg=headerStrong, fg=white, font=\bfseries},
}
Model & Raw Output & Aggregation & Dim \\
MERT-v0~\cite{li2024mert} &
  \SetCell[r=2]{c,m} $\mathbb{R}^{13 \times T \times 768}$ &
  \SetCell[r=2]{c,m} Mean over $T$; concat all $L{+}1$ layers; L2-norm &
  \SetCell[r=2]{c,m} 9{,}984 \\
MERT-v0-pub\tnote{1}~\cite{li2024mert} & & & \\
Music2Vec-v1~\cite{li2022map} & $\mathbb{R}^{13 \times T \times 768}$ & Mean over $T$; concat all $L{+}1$ layers; L2-norm & 9{,}984 \\
DAC-16k~\cite{kumar2023high} & $\{0,\ldots,1023\}^{12 \times T}$ & Histogram per codebook; concat; L2-norm & 12{,}288 \\
\end{tblr}
\begin{tablenotes}[flushleft]
\scriptsize
\item[1] MERT-v0-pub is trained exclusively on publicly licensed audio; MERT-v0 additionally incorporates a private corpus. Both share the same architecture and produce identical-dimensional embeddings.
\end{tablenotes}
\end{threeparttable}
\end{table*}

\subsection{Attribution Performance Under Different Metrics}
\label{sec:differentmetics}
\begin{figure}[h]
    \centering
    \includegraphics[width=1\linewidth]{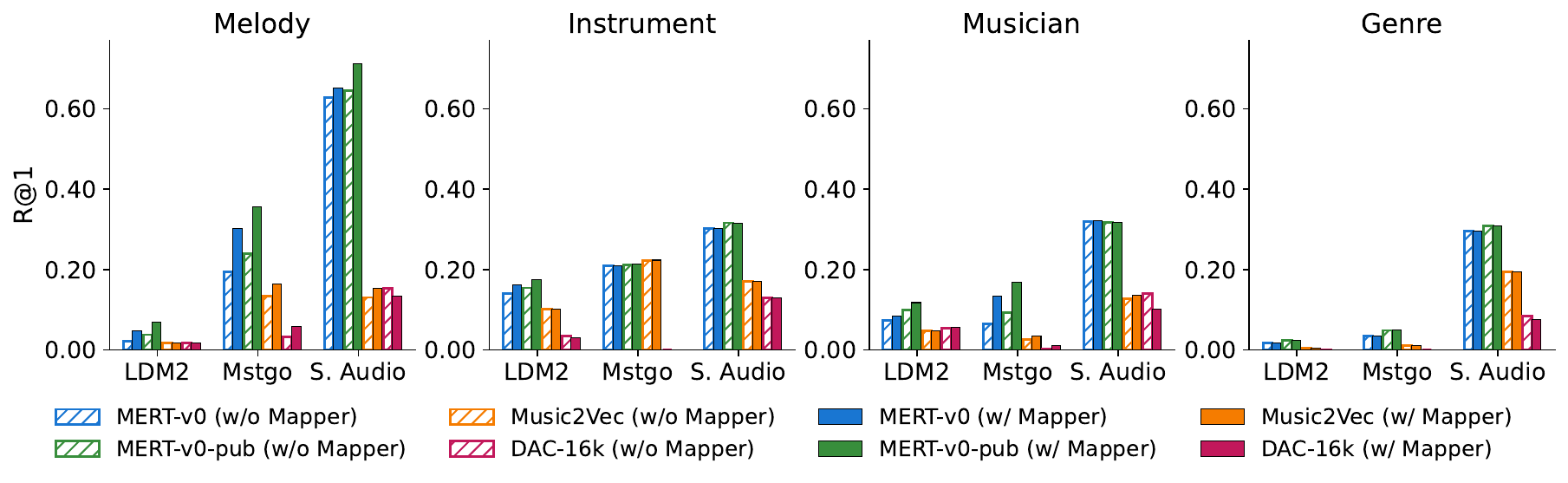}
    \caption{Attribution R@1 with and without learned mappers, using $3$ clips per attribute. The generator names are abbreviated as LDM2 (AudioLDM2), Mstgo (Mustango), and S. Audio (Stable Audio Open).}
    \label{fig:metrics-r1-3clips}
\end{figure}

\begin{figure}[h]
    \centering
    \includegraphics[width=1\linewidth]{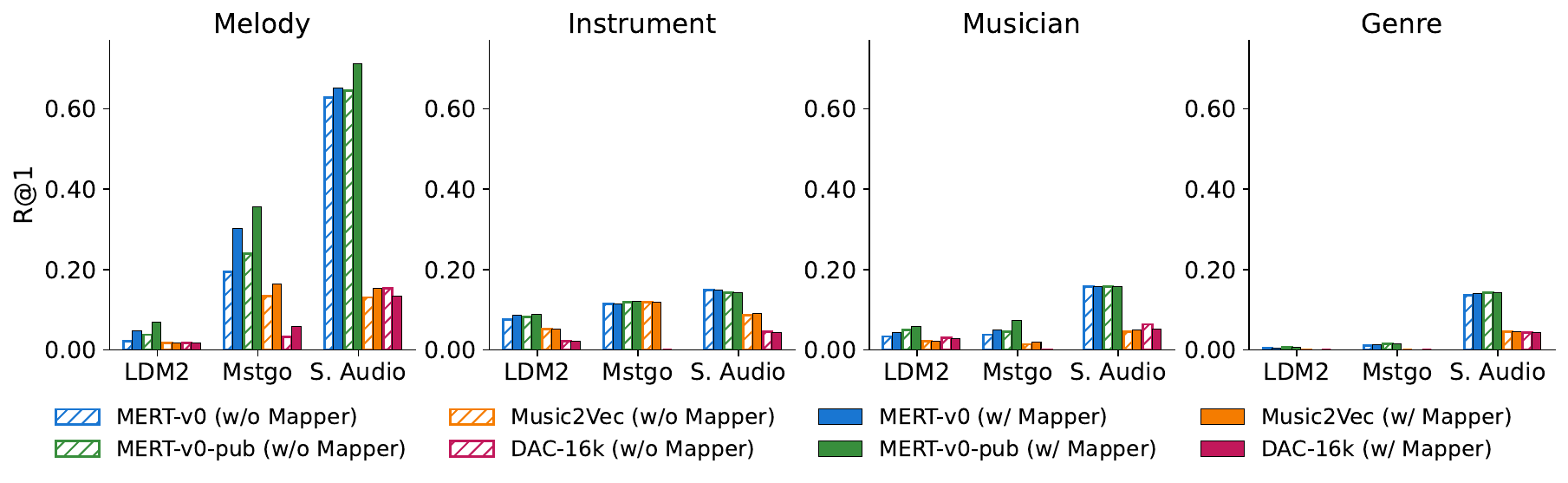}
    \caption{Attribution R@1 with and without learned mappers, using $6$ clips per attribute. The generator names are abbreviated as LDM2 (AudioLDM2), Mstgo (Mustango), and S. Audio (Stable Audio Open).}
    \label{fig:metrics-r1-6clips}
\end{figure}

\begin{figure}[h]
    \centering
    \includegraphics[width=1\linewidth]{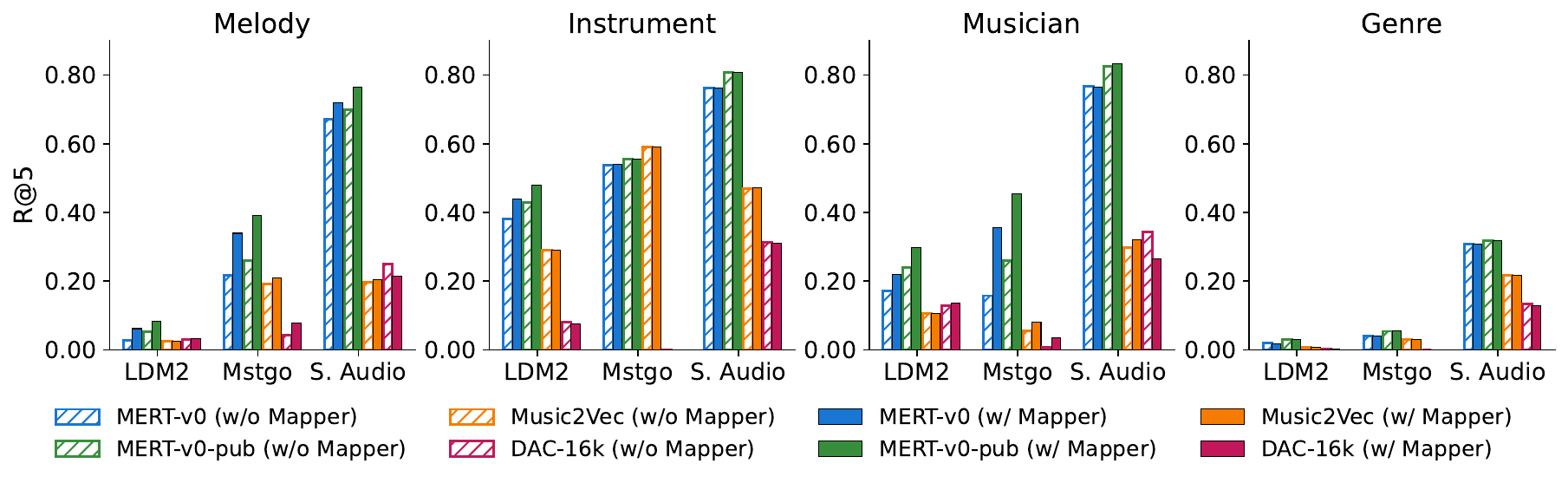}
    \caption{Attribution R@5 with and without learned mappers, using $3$ clips per attribute. The generator names are abbreviated as LDM2 (AudioLDM2), Mstgo (Mustango), and S. Audio (Stable Audio Open).}
    \label{fig:metrics-r5-3clips}
\end{figure}

\begin{figure}[h]
    \centering
    \includegraphics[width=1\linewidth]{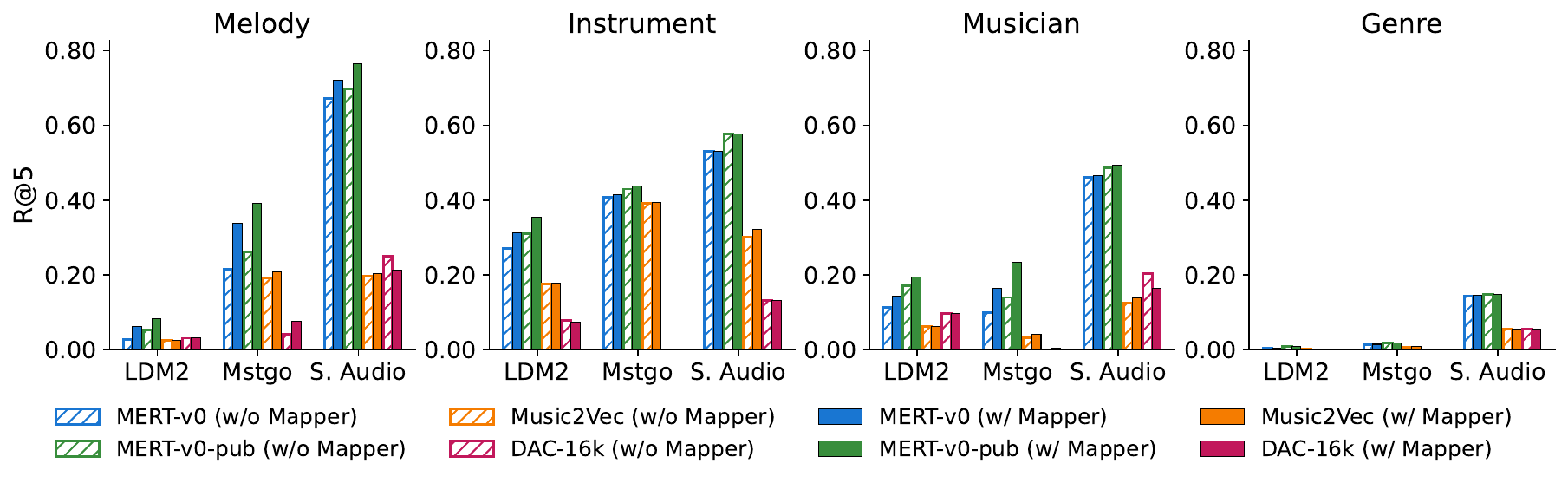}
    \caption{Attribution R@5 with and without learned mappers, using $6$ clips per attribute. The generator names are abbreviated as LDM2 (AudioLDM2), Mstgo (Mustango), and S. Audio (Stable Audio Open).}
    \label{fig:metrics-r5-6clips}
\end{figure}

\begin{figure}[h]
    \centering
    \includegraphics[width=1\linewidth]{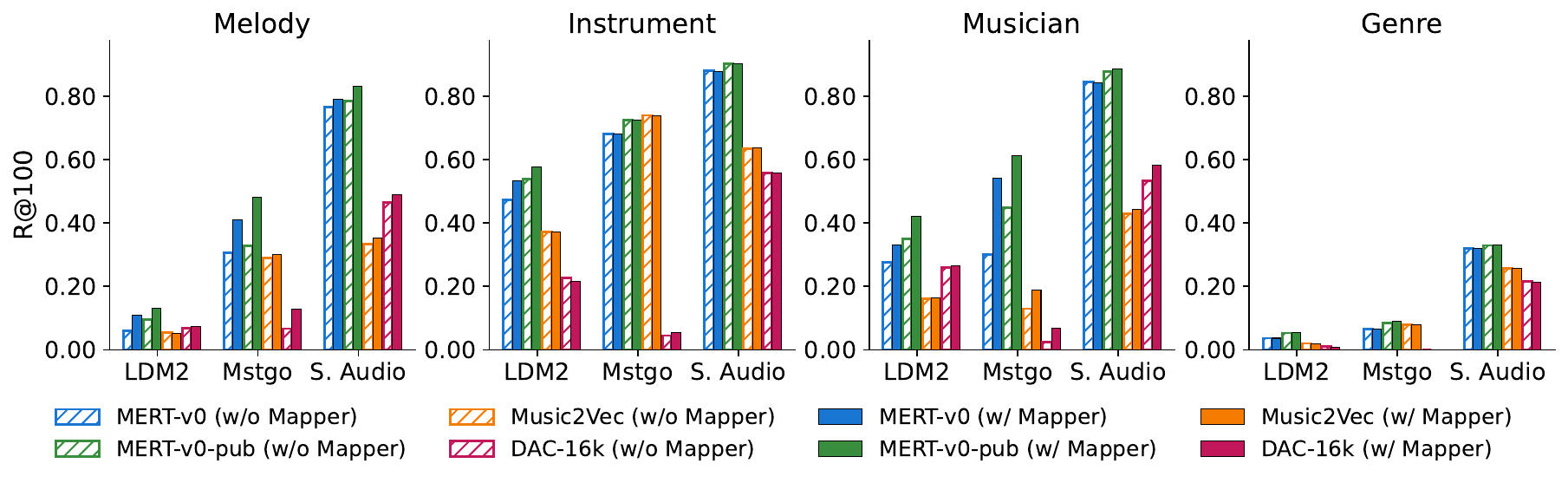}
    \caption{Attribution R@100 with and without learned mappers, using $3$ clips per attribute. The generator names are abbreviated as LDM2 (AudioLDM2), Mstgo (Mustango), and S. Audio (Stable Audio Open).}
    \label{fig:metrics-r100-3clips}
\end{figure}

\begin{figure}[h]
    \centering
    \includegraphics[width=1\linewidth]{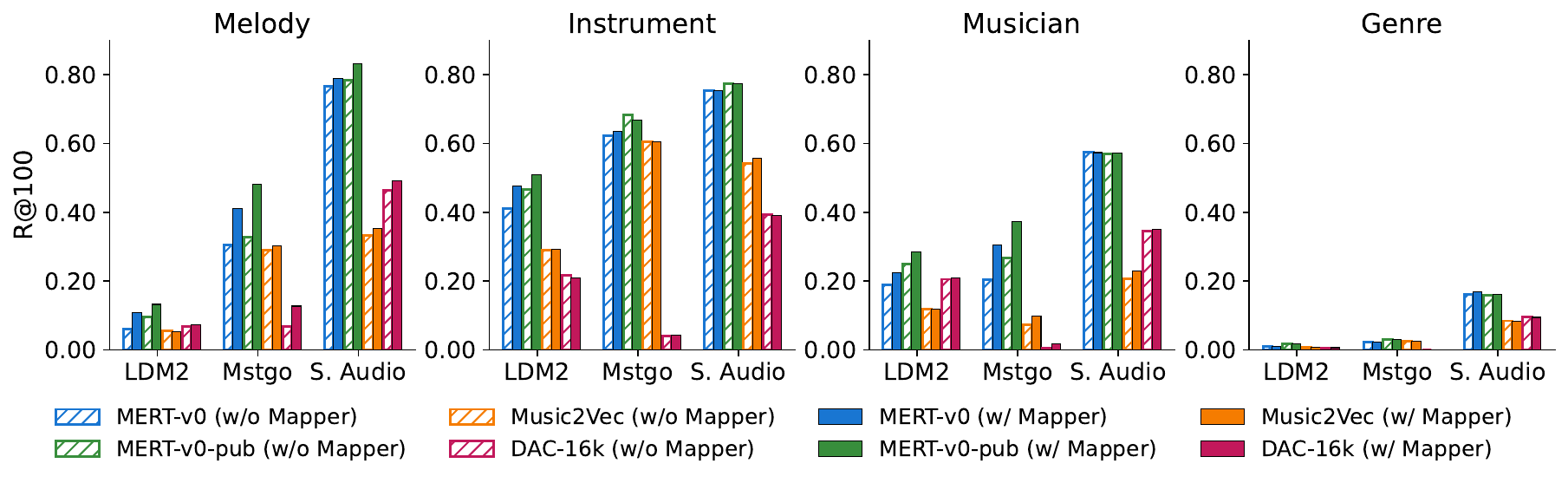}
    \caption{Attribution R@100 with and without learned mappers, using $6$ clips per attribute. The generator names are abbreviated as LDM2 (AudioLDM2), Mstgo (Mustango), and S. Audio (Stable Audio Open).}
    \label{fig:metrics-r100-6clips}
\end{figure}

\begin{figure}[h]
    \centering
    \includegraphics[width=1\linewidth]{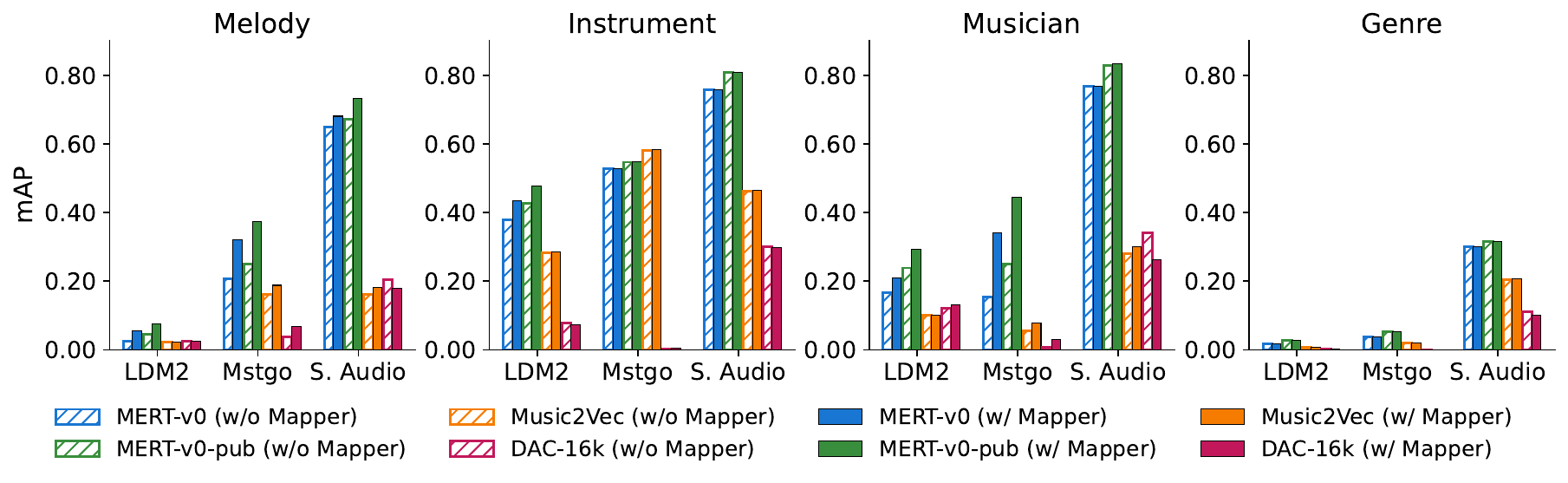}
    \caption{Attribution Mean Average Precision (MAP) with and without learned mappers, using $3$ clips per attribute. The generator names are abbreviated as LDM2 (AudioLDM2), Mstgo (Mustango), and S. Audio (Stable Audio Open).}
    \label{fig:metrics-map-3clips}
\end{figure}

\begin{figure}[h]
    \centering
    \includegraphics[width=1\linewidth]{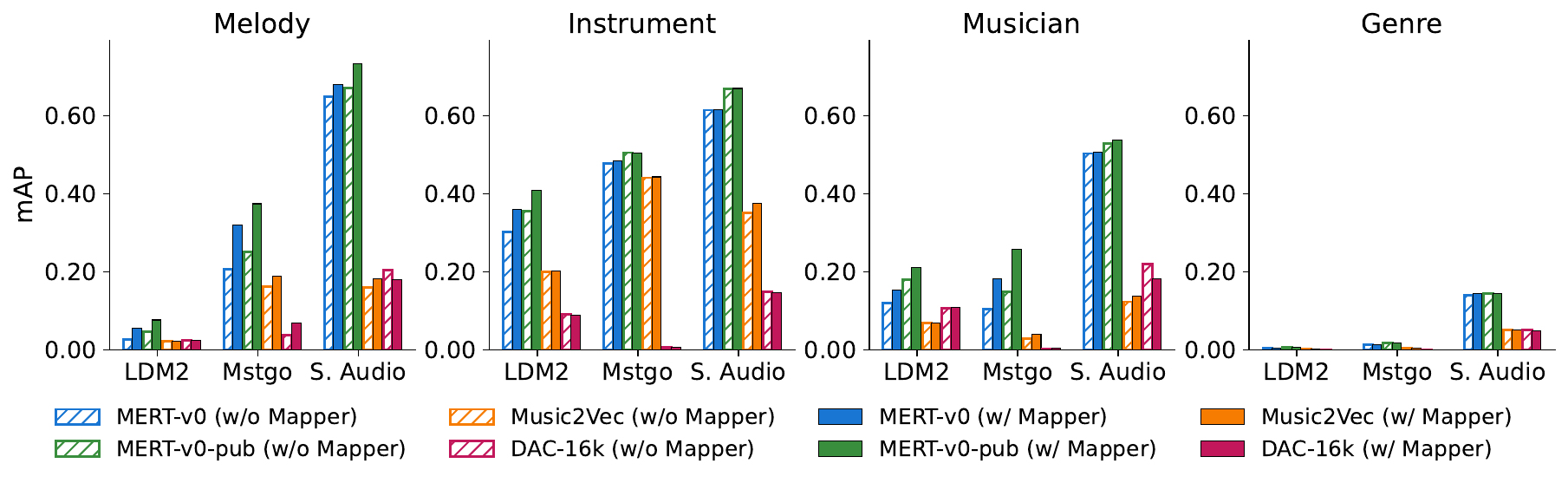}
    \caption{Attribution Mean Average Precision (MAP) with and without learned mappers, using $6$ clips per attribute. The generator names are abbreviated as LDM2 (AudioLDM2), Mstgo (Mustango), and S. Audio (Stable Audio Open).}
    \label{fig:metrics-map-6clips}
\end{figure}

We report attribution performance under additional metrics, including R@1, R@5, R@100, and Mean Average Precision (MAP), for the 3-clip setting and for the 6-clip setting of the non-melody attribution sets. The overall trends are consistent with those observed under R@10 in the main text: MERT-based encoders consistently outperform Music2Vec-v1 and DAC-16k across all metrics, Stable Audio Open yields the highest scores, and learned mappers provide the most benefit for MERT-based encoders on Mustango. As expected, R@100 produces uniformly higher scores than R@10, while R@1 and MAP are more stringent and reveal larger performance gaps between encoders. The relative ordering across encoders, generators, and attribution sets remains stable across all metrics, confirming that the conclusions drawn from R@10 in the main text generalize to other retrieval metrics. Results are shown in Figures~\ref{fig:metrics-r1-3clips},~\ref{fig:metrics-r1-6clips},~\ref{fig:metrics-r5-3clips},~\ref{fig:metrics-r5-6clips},~\ref{fig:metrics-r100-3clips},~\ref{fig:metrics-r100-6clips},~\ref{fig:metrics-map-3clips}, and~\ref{fig:metrics-map-6clips}.

\subsection{Implementation Details}
\subsubsection{Finetuning Details for Text-to-Music Models}
\label{sec:detailsfort2m}
We finetune each text-to-music model per attribute using a DreamBooth-style procedure~\cite{ruiz2023dreambooth}. Each attribute is represented by a rare identifier ``sks'' prepended to the class name (e.g., ``sks musician''), and the model is trained to associate this identifier with the reference music clips. We use a constant learning rate of $4 \times 10^{-6}$ with the AdamW optimizer for AudioLDM2 and Mustango, and the Adam optimizer for Stable Audio. For all three generators, we finetune only the denoising U-Net, keeping all other components such as text encoders, VAE, and any music-structure embeddings frozen. The number of training steps and batch size depend on the number of reference clips: for melody attributes (1 clip) we train for 300 steps with batch size 1; for 3 clips we train for 1{,}000 steps with batch size 3; for 6 clips we train for 1{,}500 steps with batch size 4.

\subsubsection{Training Details for Feature Mappers}
\label{sec:detailsformappers}
We initialize the feature mapper with a zero delta weight matrix and zero bias, so that the mapper reduces to the identity at the start of training. We use Adam optimizer with $\beta_1=0.9$, $\beta_2=0.999$ and an initial learning rate of $10^{-3}$, decayed with a cosine annealing schedule over the full training horizon. The batch size is 32. We train each mapper for up to 200 epochs with early stopping (patience 20 on validation R@10). Each epoch involves sampling one ground truth--query pair per attribute from the train split; within each step, the batch is a random subset of these pairs drawn without replacement. The training objective is the symmetric NT-Xent loss (temperature $\beta=1$) between the mapped ground truth and query embeddings, plus a Frobenius-norm regularization term on the delta weight matrix (coefficient $\lambda=0.05$).

\subsubsection{Computational Resource Usage}
\label{compuationalresource}
All experiments were conducted on NVIDIA A100 80\,GB GPUs, requiring approximately 637 GPU-hours across all stages. This includes attribute-level fine-tuning of the text-to-music generators, generation of the full audio gallery, embedding extraction with four music encoders, and mapper training. Most of the computation was spent on generator fine-tuning and sample generation, while embedding extraction and mapper training incurred relatively small additional cost.

\subsection{Attribution Performance Under Different Prompt Types per Generator}
\label{sec:allgenerator}
\begin{figure}[h]
    \centering
    \includegraphics[width=1\linewidth]{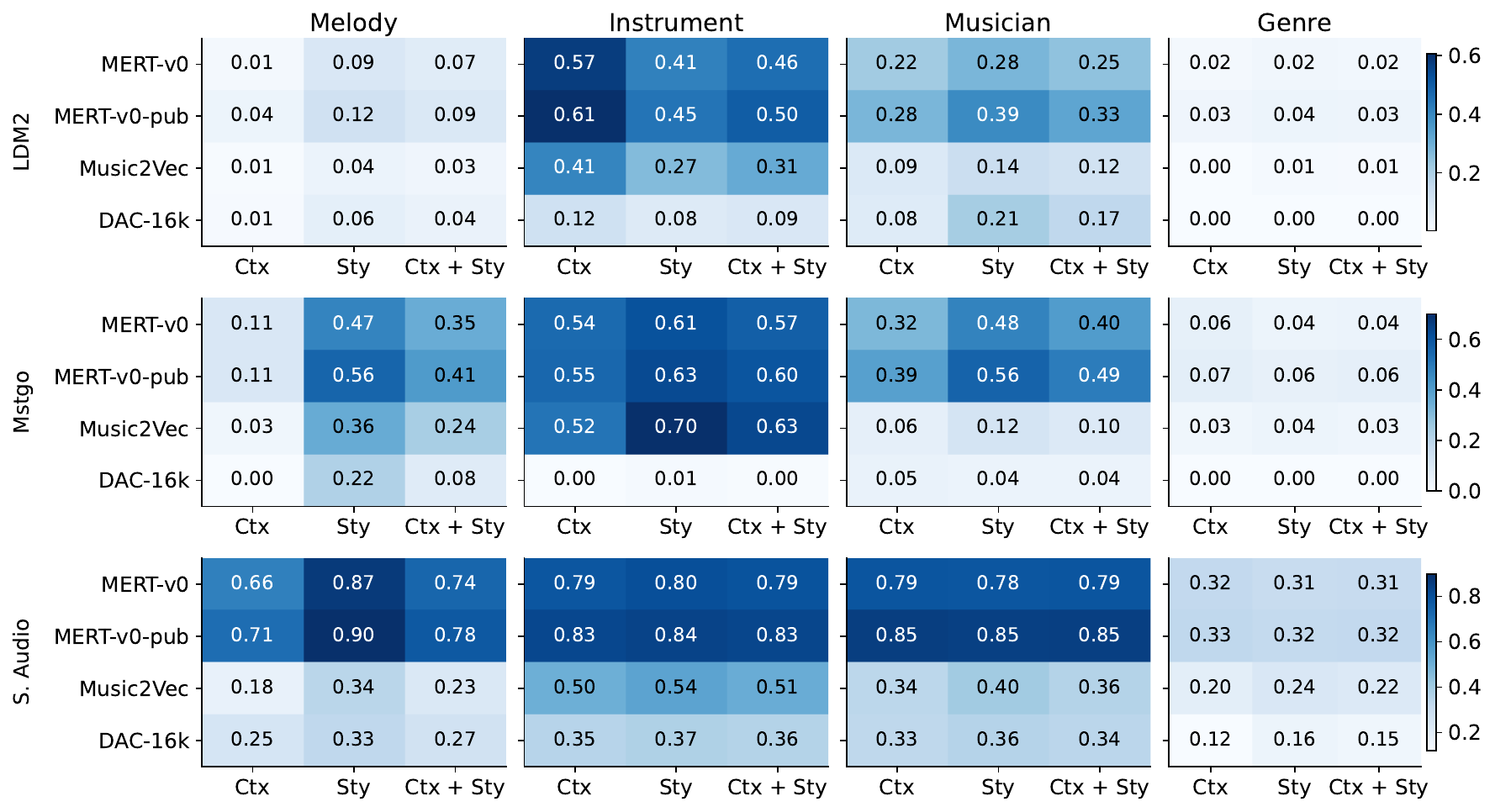}
    \caption{Attribution performance under different prompt types broken down by generator, using 3 clips per attribute. Each row corresponds to one generator; each panel corresponds to one attribution set.}
    \label{fig:multiprompt-pergen-3clips}
\end{figure}

\begin{figure}[h]
    \centering
    \includegraphics[width=1\linewidth]{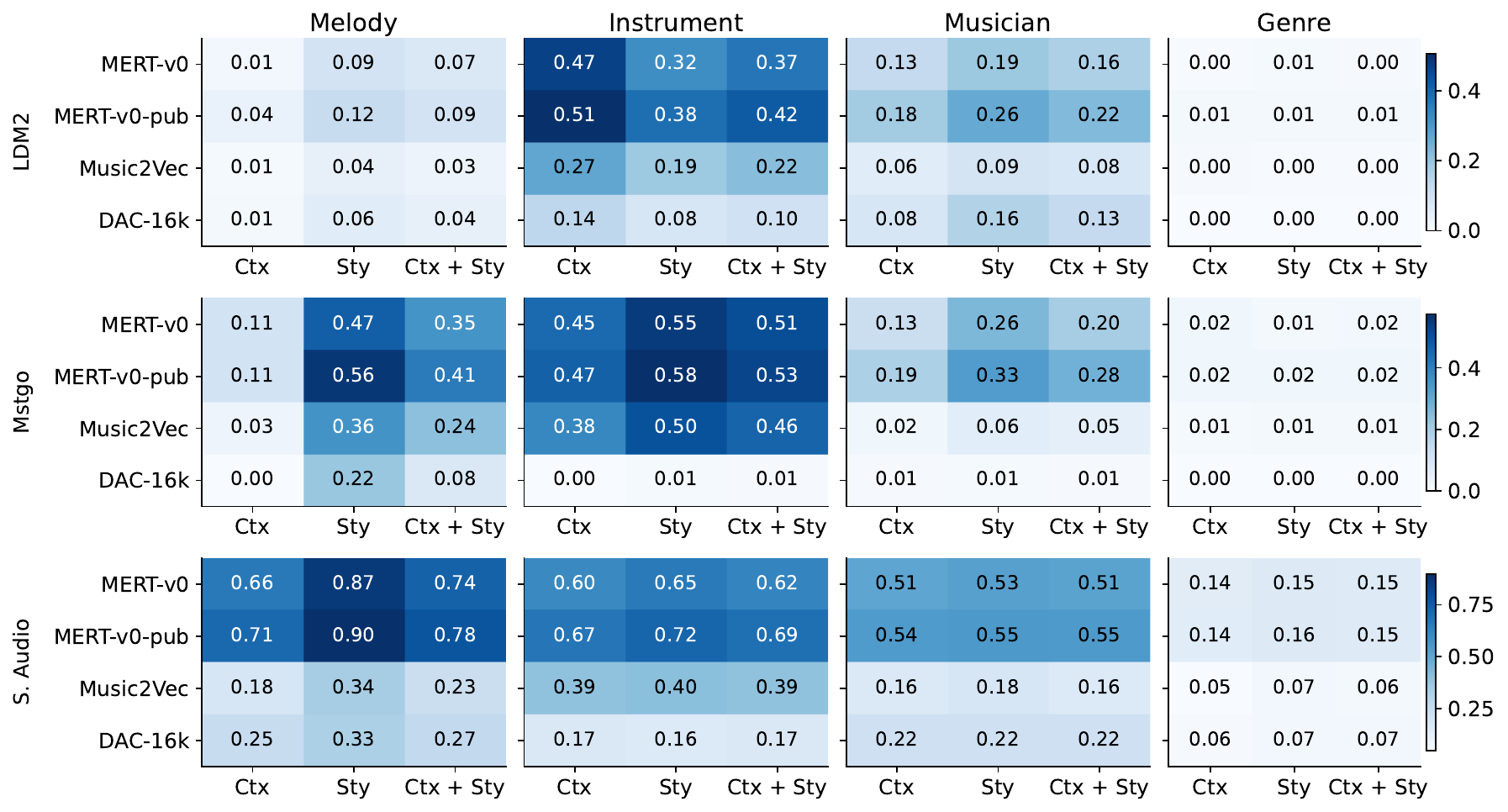}
    \caption{Attribution performance under different prompt types broken down by generator, using 6 clips per attribute. Each row corresponds to one generator; each panel corresponds to one attribution set.}
    \label{fig:multiprompt-pergen-6clips}
\end{figure}

We report the per-generator breakdown of prompt type experiments for both the 3-clip and 6-clip settings. The style advantage observed in the main text holds across all three generators, though the magnitude varies. On Stable Audio Open, the gap between style and contextual prompts is smaller, as the generator's tendency to memorize fine-tuning clips already produces consistent outputs regardless of prompt type. On Mustango and AudioLDM2, the gap is more pronounced, particularly for Melody-Centric, where contextual prompts consistently yield near-zero R@10 while style prompts produce substantially higher scores. Results are shown in Figures~\ref{fig:multiprompt-pergen-3clips} and~\ref{fig:multiprompt-pergen-6clips}.


\subsection{Open Access to Code and Data}
\label{Openaccess}
We release the \methodName datasets, and source code to support reproducibility and future research. The benchmark dataset is available at \url{https://huggingface.co/datasets/AnonymousAuthorsssss/TrueMuse}. 

The code for dataset construction, attribution evaluation, and experimental reproduction is available at \url{https://github.com/ZeliuFoo/TrueMuse}.

%%%%%%%%%%%%%%%%%%%%%%%%%%%%%%%%%%%%%%%%%%%%%%%%%%%%%%%%%%%%

\newpage
\clearpage

\end{document}